\documentclass[final]{cvpr}

\usepackage{times}
\usepackage{epsfig}
\usepackage{graphicx}
\usepackage{amsmath}
\usepackage{amssymb}
\usepackage{multirow}
\usepackage{comment}
\usepackage{subfig}
\usepackage{xcolor,booktabs,amsthm}

\usepackage[pagebackref=true,breaklinks=true,colorlinks,bookmarks=false]{hyperref}

\def\cvprPaperID{5073} 
\def\confYear{CVPR 2021}

\begin{document}

\title{Adjusting Bias in Long Range Stereo Matching: A semantics guided approach}

\author{WeiQin Chuah, Ruwan Tennakoon, Reza Hoseinnezhad
        and~Alireza~Bab-Hadiashar\\
RMIT University, Australia\\
{\tt\small wei.qin.chuah@student.rmit.edu.au, \{ruwan.tennakoon,rezah,abh\}@rmit.edu.au}
\and
David Suter\\
Edith Cowan University~(ECU), Australia\\
{\tt\small d.suter@ecu.edu.au}
}

\maketitle

\begin{abstract}
Stereo vision generally involves the computation of pixel correspondences and estimation of disparities between rectified image pairs. In many applications, including simultaneous localization and mapping (SLAM) and 3D object detection, the disparities are primarily needed to calculate depth values and the accuracy of depth estimation is often more compelling than disparity estimation. The accuracy of disparity estimation, however, does not directly translate to the accuracy of depth estimation, especially for faraway objects. In the context of learning-based stereo systems, this is largely due to biases imposed by the choices of the disparity-based loss function and the training data. Consequently, the learning algorithms often produce unreliable depth estimates of foreground objects, particularly at large distances~($>50$m). To resolve this issue, we first analyze the effect of those biases and then propose a pair of novel depth-based loss functions for foreground and background, separately. These loss functions are tunable and can balance the inherent bias of the stereo learning algorithms. The efficacy of our solution is demonstrated by an extensive set of experiments, which are benchmarked against state of the art. We show on KITTI~2015 benchmark that our proposed solution yields substantial improvements in disparity and depth estimation, particularly for objects located at distances beyond 50 meters, outperforming the previous state of the art by $10\%$. Our code is available at: \textit{\{link removed to maintain anonymity\}}. 

\end{abstract}

\section{Introduction} \label{sec:intro}

\begin{figure}[th!] 
\centering
		\includegraphics[width=0.45\textwidth]{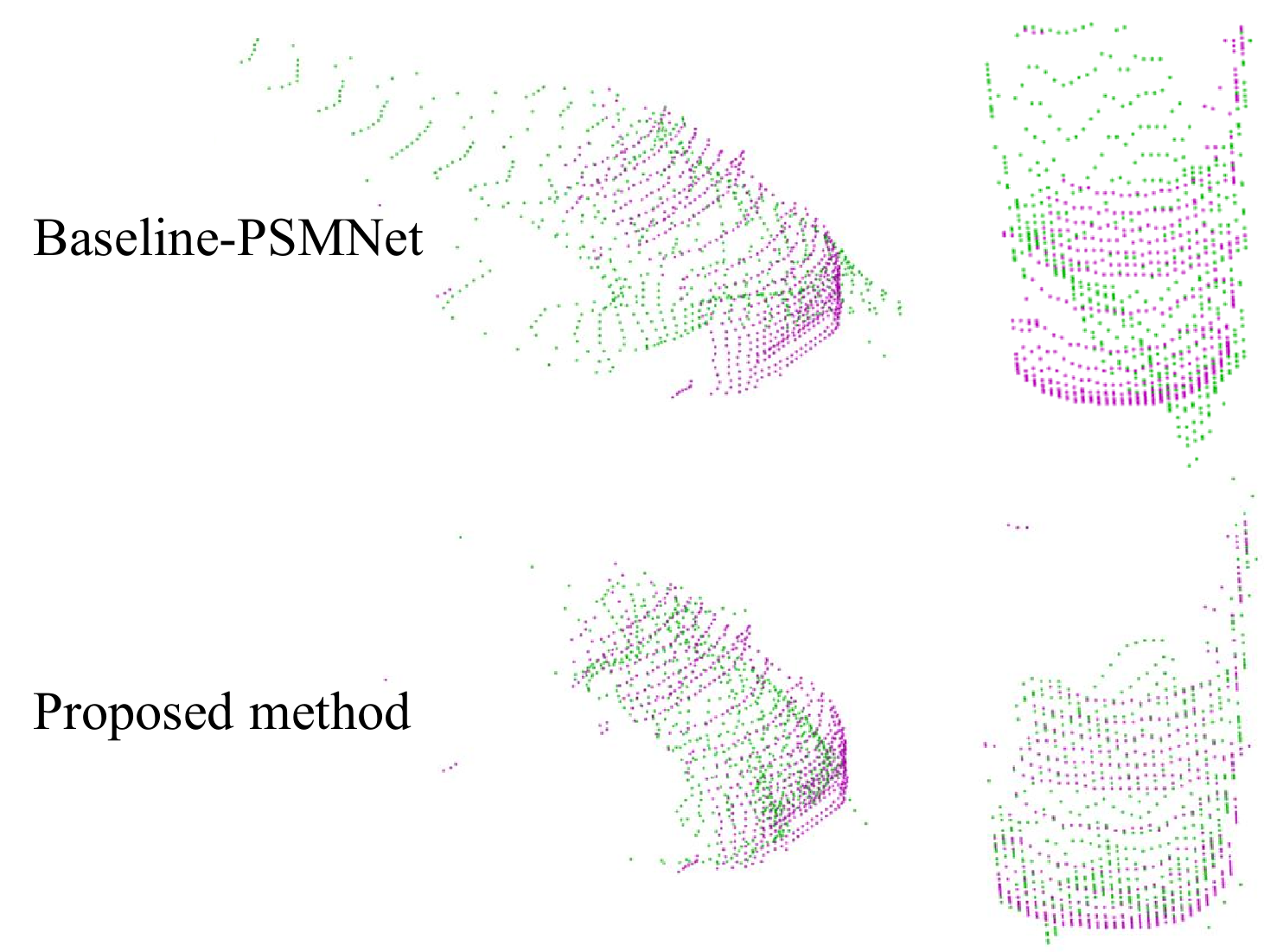}
		
	\caption{An example of back-projected 3D points and estimated disparity of an object at 43 meters from the camera (image $000038\_10$ of KITTI 2015 dataset). Ground truth points are plotted in magenta while predicted points are shown in green. Top and bottom rows show the result of the baseline method~\cite{chang2018pyramid} and our proposed method. The picture shows that the proposed method significantly improves the depth estimation of a far object.}
	\label{fig:3d_improve}
\end{figure}

Accurate depth perception is critical in applications, such as autonomous driving, robot navigation, and 3D reconstruction. This can be accomplished by estimating pixel correspondences and disparities between rectified image pairs, which is known as stereo matching. 
In recent years, there have been significant interests in developing learning-based stereo matching and disparity estimation algorithms. Most of these works have focused on improving the performance of disparity estimation by designing more effective network architectures~\cite{chang2018pyramid, guo2019group, cheng2019learning, zhang2019ga} or loss functions with stronger training signals~\cite{zhang2019adaptive}. Despite significant progress, the combination of existing disparity-based loss functions and the commonly used training data biases the models towards emphasising more on near-by objects and background areas at the expense of farther objects~\cite{you2019pseudo, qian2020end} . This effect can be detrimental, especially in autonomous navigation-related applications.

Before discussing the identified problems in learning-based stereo disparity estimation algorithms, it maybe useful to clarify the key terms such as: (1)~\textit{bias}, (2)~\textit{near, middle distance, and far} and (3)~\textit{foreground\,/\,background}. In statistical literature, bias has a precise meaning, and for an estimator it is defined as the difference between the expected value of the estimated parameter and its true value~\cite{Goodfellow-et-al-2016}. However, in machine learning literature, the term bias is also loosely used to indicate ``putting too much or too little weight'' on an attribute compared to some natural uniform measure, which in turn can introduce a bias in the statistical sense. The three terms (\textit{near, middle distance, and far}) have obvious meanings at least in terms of relative order. In most cases, these terms are associated with some thresholds that are somewhat context or application dependent. Thus, it is naive to expect any fixed setting of these thresholds will be universally useful. In this work, we selected a reasonable threshold that is appropriate for the common datasets that are designed for different autonomous navigation scenarios. While the terms (\textit{background\,/\,foreground}) may intuitively suggest of far and near objects, in this paper, we refer to them as ``certain objects of interest in the scene"~(foreground) and the others~(background). In our analysis, we have purposely chosen objects that are semantically important in autonomous navigation tasks.

\begin{figure}[t]
\centering
	\includegraphics[width=0.33\textwidth]{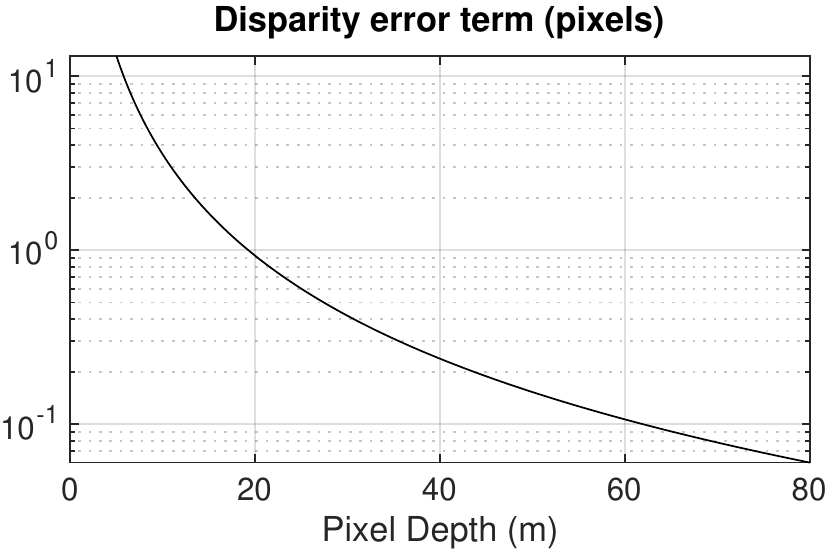}
	\caption{Variations of the disparity loss term (a pixel disparity error term) with the depth of the pixel, corresponding to 1 meter depth error.}
	\label{fig:graph1}
\end{figure}

In learning-based stereo disparity estimation algorithms, the implemented loss functions are usually disparity-based~\cite{mayer2016large, chang2018pyramid, guo2019group, cheng2019learning, zhang2019ga, zhang2019adaptive, xu2020aanet}. 
However, in many applications, particularly in autonomous navigation (e.g.~simultaneous localization and mapping (SLAM) and 3D object detection), depth values are often preferred and are derived from the estimated disparities. 
The error in disparity estimation does not translate directly to the error in depth estimation as the disparity-based loss functions does not uniformly penalize errors at different distances~\cite{you2019pseudo}. 
As illustrated in Figure~\ref{fig:graph1}, one metre error at any distance translates to disparity-based losses that would penalize near-by objects ($\leq$ 20\,m) much higher than far objects. For instance, for KITTI~2015~\cite{Menze2018JPRS} where the focal length and baseline are 721 pixels and 0.54\,m, one meter depth error at 5 meter distance corresponds to $13$ pixel disparity error while it only corresponds to $0.6$ pixel error at 25 meter distance.

\begin{figure}[t]
\centering
	\includegraphics[width=\columnwidth]{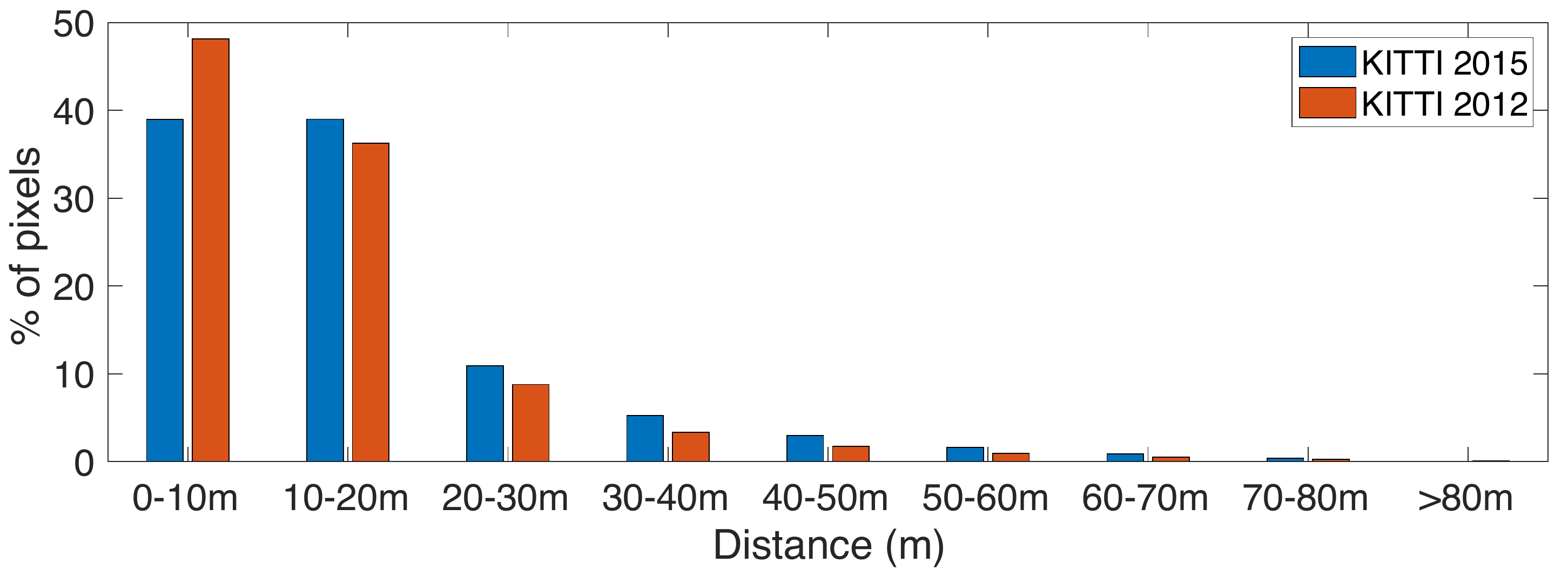}
	\caption{Distribution of pixels associated with different distances in KITTI~2012 and KITTI~2015 stereo dataset.}
	\label{fig:data_distribution}
\end{figure}

Furthermore, the front-view of most driving scenes are primarily occupied by (1)~foreground that are close to the vehicle or (2)~background. For example, in KITTI~2015, approximately $65\%$ of total pixels in an image are background and have depth value $\leq 20$\,m. More severely, only $2\%$ of the total pixels are foreground and have depth value $>20$\,m. The details of pixel data distribution are included in Table~\ref{tab:data} and Figure~\ref{fig:data_distribution}. 

Therefore, final solutions of the learning algorithms, utilizing the disparity-based loss functions and the mentioned training data will results in estimates that are biased towards background and are somewhat "blinded" to the foreground~(especially those that are positioned at farther distances). As mentioned, these conditions are detrimental for autonomous navigation related applications. For example, as illustrated in Fig. \ref{fig:3d_improve}, the depth estimates resulted by a state-of-the-art method~\cite{chang2018pyramid} without addressing the bias issues can be extremely unreliable for far objects. To address these issues, we propose to adjust the bias of the conventional stereo learning algorithms to consider far and foreground objects more favourably without losing the sight of the close-by objects and background areas. 

In short, we propose to regularize the bias of learning-based stereo disparity estimation algorithms by adjusting the loss function based on the notion of relevant depth-ranges and scene contents. More specifically, we use a depth based loss function which is bisected into foreground and background segments using an off-the-shelf object detector to balance the bias between the two classes.

Although, shifting the bias or preference of a stereo disparity estimation algorithm from disparity to depth may intuitively suggest some deterioration in the performance for the near-by objects and background areas, our results demonstrate otherwise. Our results illustrate that the proposed loss functions can significantly improve the overall accuracy of disparity and depth estimates at all distances, outperforming the previous state-of-the-art long range stereo depth estimation technique that utilizes  accurate LiDAR (active) measurements for depth refinements~\cite{you2019pseudo}. We also show that a simple change of the loss function in the PSMNet~\cite{chang2018pyramid} method can improve its rank in the KITTI benchmark from $116^{\text{th}}$ to $64^{\text{th}}$ place (recorded on $8^{\text{th}}$ of November 2020). 



\section{Related Works} \label{sec:related-works}
\subsection{Learning-Based Stereo Matching Networks}
Recent works~\cite{zbontar2016stereo, luo2016efficient, chang2018pyramid, zhang2019ga, liang2018learning, guo2019group} have shown that stereo matching using deep features has significant performance boost over traditional hand-crafted features like SIFT~\cite{lowe2004distinctive} and ORB~\cite{rublee2011orb} features. Existing end-to-end stereo matching networks utilized CNN to (1) extract deep representation from input stereo images, (2) cost volume aggregations and (3) cost volume refinement. 

\begin{table}[t]
\footnotesize
\centering
\begin{tabular}{lrrr}
\toprule
           & $\leq$ 20\,m & $>$20\,m  & Total    \\ \toprule
Foreground & 14.90\%   & 1.91\%  & 16.81\%  \\ \midrule
Background & 64.79\%   & 18.40\% & 83.19\%  \\\midrule
Total      & 79.69\%   & 20.31\% & 100.00\% \\ \bottomrule
\end{tabular}%
\caption{Pixel data distribution of KITTI 2015 dataset.} 
\label{tab:data}
\end{table}

In terms of taxonomy, end-to-end stereo matching networks can be classified into two categories: (1) correlation-based and (2) shifted concatenation-based cost volume construction methods. The correlation-based networks consist of stacked 2D CNNs layers and have significantly lower processing time due to the high efficiency of 2D convolution~\cite{mayer2016large,xu2020aanet,liang2018learning, pang2017cascade}. The concatenation based networks consist of a combination of 2D CNNs for feature extraction and 3D CNNs for cost volume aggregation and refinement~\cite{kendall2017end,zhang2019ga,chang2018pyramid}. An interesting exception is the idea of group-wise correlation-based cost volume construction that was proposed to preserve information loss of full correlation~\cite{guo2019group}.
In terms of performance, shifted concatenation-based networks with 3D CNNs layers often outperform correlation-based networks with 2D CNNs layers by a large margin on popular benchmarks (e.g., SceneFlow~\cite{mayer2016large}, KITTI~\cite{Menze2018JPRS}). 

To close the performance gaps between the correlation and the shifted concatenation methods, several prior works proposed to include context information such as edges~\cite{song2018edgestereo} and semantics~\cite{yang2018segstereo} into the network. Moreover, Xu~\etal~\cite{xu2020aanet} proposed AANet which consists of a new sparse points-based representation for intra-scale aggregation and adaptive multi-scale cross aggregation modules using 2D CNNs layers. As a results, AANet has comparable performance as the shifted concatenated methods but with real-time inference speed.

Despite all the advances in learning-based stereo matching network, the relationship between disparity and depth is rarely discussed. While current state-of-the-art stereo matching network is capable of performing disparity estimation with high accuracy, the accuracy in disparity estimation does not translate directly to the accuracy in depth estimation, especially for objects that are far away from the camera, as mentioned in Section~\ref{sec:intro}. Different from existing methods, we turn our attention to this problem and propose a simple yet effective solution, which allows our method to improve the performance of both disparity and depth estimation, particularly for far away foreground objects. 

\subsection{Depth estimation and 3D object detection}

Accurate depth information of moving (foreground) objects such as pedestrians, transportation vehicles and cyclist is important in downstream applications such as 3D object detection and autonomous navigation. There are several works that concentrate on stereo depth estimation for 3D object detection. For instance, Pon~\etal~\cite{pon2019object} proposed a stereo matching network that focuses on objects of interest while neglecting the background. 
Qian~\etal~\cite{qian2020end} proposed to combine stereo matching and 3D object detection networks into a single pipeline by designing a novel differentiable module to convert predicted depth map to pseudo-LiDAR~\cite{wang2019pseudo}. They used the same stereo matching network proposed in~\cite{you2019pseudo}. 

Although existing methods can achieve impressive results for 3D object detection from RGB images, the performance exacerbates as the distance increases due to the factors discussed in Section~\ref{sec:intro}. In this context, You~\etal~\cite{you2019pseudo} proposed SDN aiming to improve the long range depth estimation by converting a disparity-based stereo matching network~\cite{chang2018pyramid} into a depth-based stereo matching one. The proposed network converts the disparity cost volume to depth cost volume thus regressing depth value for each pixel instead of disparity. They further proposed a depth propagation algorithm which fuses extremely sparse (4-beam) LiDAR to rectify the initial depth estimation.

In contrast, we aim to improve the performance of depth estimation of a stereo matching network by adjusting the bias in training loss function and selected dataset. We propose to carefully balance the training signals, preventing any over-emphasis on backgrounds or close objects, as mentioned in Section~\ref{sec:intro}. As a result, the proposed method achieves significant improvement in disparity and depth estimation, particularly for distant objects, over the baseline method. More importantly, our results are on par with the mentioned prior work SDN, without using any additional information such as sparse LiDAR data points.

\section{Proposed Method} \label{sec:method}
In this section, we will discuss our proposed loss functions and the overall framework to address the challenges of existing bias for long range stereo depth estimation as explained in Section \ref{sec:intro}. The results of our experimental and ablation studies are presented in Section \ref{sec:experiment}. 

\subsection{Loss Function}
As the performance of supervised learning neural network largely depends on its loss function, it is crucial to select the appropriate loss function carefully. Also, a properly designed loss function can mitigate the adverse effect of bias (such as data imbalance, class imbalance) in the training dataset, and therefore improve the overall performance of the trained model~\cite{Lin_2017_ICCV}. We show that using solely disparity-based smooth $L_1$ loss function would cause the trained model to overfit\footnotemark~to nearby objects and background areas, resulting in worse accuracy for long-range depth estimation. This is intolerable in certain autonomous navigation applications. To address this issue, we propose to redesign the loss function by including foreground and background specific depth-based loss functions.
\footnotetext{We define \textit{overfit} as: a network is learning to fit accurately at a certain part of the data (dominant data points) with the expenses of lower accuracy for other parts. This is different from the general definition of \textit{overfit} in machine learning: a network is learning to also fit the noise in the training data, and negatively impacts the network capability to generalise to new data.}

\vfill
\noindent \textbf{Disparity-based loss function: }
The disparity-based loss function is implemented to enable the stereo matching network to learn the regression of disparity for each pixel. The smooth $L1$ disparity-based loss function is defined as:
\begin{equation}
	\mathcal{L}_{disp} = \frac{1}{N}\sum_{i=1}^{N} \text{smooth}_{L_1}(D_i-\hat{D}_i),\label{eq:dispLossFunction}
\end{equation} 
in which
\[\text{smooth}_{L_1}(x)=   
\left\{
\begin{array}{ll}
      0.5x^2 & \text{if}\ |x|<1 \\
      |x|-0.5, & \text{otherwise} \\
\end{array} 
\right. \]
where $\hat{D}_i$ and $D_i$ are the predicted disparity and its ground truth disparity label for pixel $i$, and $N$ is the number of valid pixels. 

As it was discussed in Section~\ref{sec:intro} and shown in Fig.~\ref{fig:graph1}, each error term in the disparity-based loss function is related to its associated depth error with a heavy bias towards the pixel depth itself. Indeed, for a typical depth error, the pixel on far objects would have tiny disparity error and therefore a tiny effect on the total loss. That is why the disparity-based loss function given by Eq.~\eqref{eq:dispLossFunction} is heavily biased and its minimization would only lead to accurate depth estimates for pixels of nearby objects.

\vfill 
\noindent \textbf{Depth-based loss function: }
Disparity-based loss functions are often designed to penalize the errors in disparity of close by objects while remaining lenient with errors of objects that are far from the camera. As it was explained earlier, this generates bias towards nearby objects. To address this issue, we include a depth-based loss function, which is similar to the loss function implemented in the Stereo Depth Network~(SDN)~\cite{you2019pseudo}. 

However, in contrast to the SDN, the disparity cost volume will not be converted to depth cost volume within the network. We instead propose to convert the predicted disparity map to depth map to ensure our proposed network does not regress depth value. As our aim is to build a passive system, unlike SDN, we do not include laser measurements to refine our results and use the same network structure as the PSMNet~\cite{chang2018pyramid}. The predicted dense disparity map, $\hat{D}$, as well as its corresponding ground truth, $D$, are converted to depth map, $\hat{Z}$ and $Z$. Using those, the depth loss function is defined as follows:
\begin{equation}
\centering
	\mathcal{L}_{depth} = \frac{1}{N}\sum_{i=1}^{N} \text{smooth}_{L_1}(Z_i-\hat{Z}_i).\label{eq:depthLossFunction}
\end{equation} 

However, replacing disparity-based with the depth-based loss function comes at the price of having less accuracy for close by objects ($\leq10$m)~\cite{you2019pseudo}. Therefore, we propose to include both disparity-based and depth-based loss functions. The depth-based loss function regularizes the learning to avoid solution that is over-fitted to the close distance pixels and vice versa. By combining the merits of both loss functions, the overall framework would be able to predict accurate disparity\,/\,depth for objects at a wide range of distances. 

\begin{figure}
	\centering
		\includegraphics[width=0.44\textwidth]{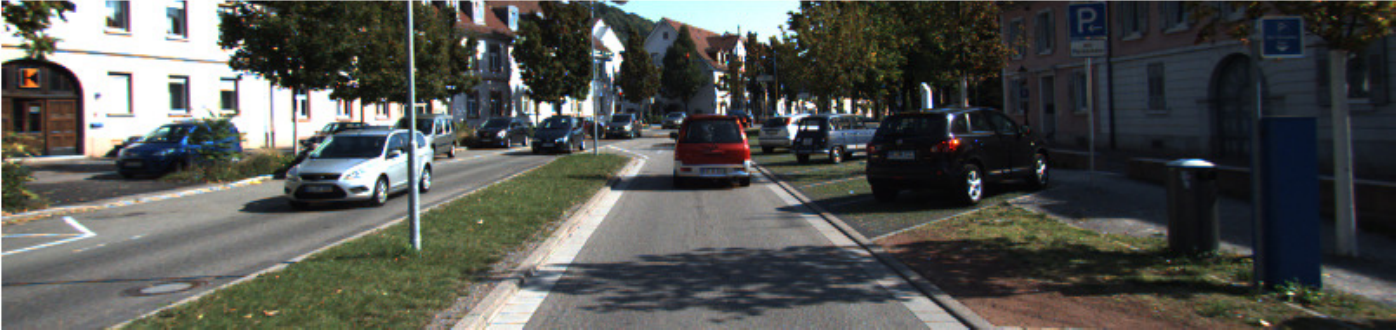}
		\includegraphics[width=0.44\textwidth]{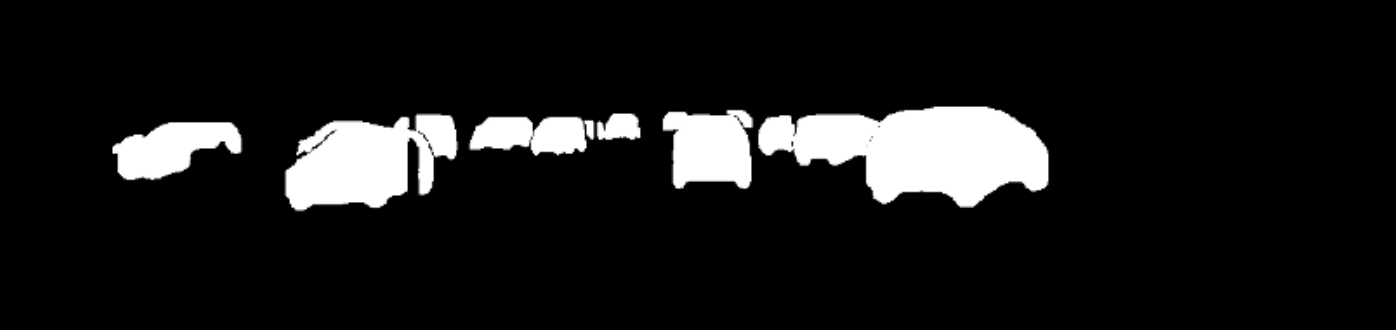}
	\caption{An example of object mask generated using a pre-trained Mask-RCNN on left image sequence $000123\_10$ of KITTI 2015 stereo dataset.}
	\label{fig:mask}
\end{figure}

\noindent \textbf{Weighted foreground and background loss functions: }
To balance the bias between foreground and background, due to class imbalance in training datasets, we propose to bisect the included depth-based loss function into two terms (foreground and background specific). These terms will be weighted accordingly and the weighting policy will be explained later. We employ Mask R-CNN~\cite{he2017mask} pre-trained on CityScapes dataset~\cite{Cordts2016Cityscapes} to extract the foreground from the background by performing foreground object segmentation. An example of the segmented foreground object masks is shown in Fig.~\ref{fig:mask}.

For simplicity, we only considered transportation vehicles including cars, trucks, vans, buses, bicycles and motorcycles as foreground objects. However, this idea can easily be extended to include other object types. We then combine the object masks and the depth loss function, expressed in Eq.~(\ref{eq:depthLossFunction}), to obtain two new loss functions that are defined as:
\begin{eqnarray}
	\mathcal{L}_{depth}^{fg} & = & 
	\frac{\sum_{i=1}^{N} \left(\text{smooth}_{L_1}(Z_i-\hat{Z}_i) \cdot\mathcal{B}_i\right)}
	{\sum_{i=1}^{N} \mathcal{B}_i}
	\\
	\mathcal{L}_{depth}^{bg} & = & 
	\frac{\sum_{i=1}^{N} \left(\text{smooth}_{L_1}(Z_i-\hat{Z}_i) \cdot(1-\mathcal{B}_i)\right)}
	{\sum_{i=1}^{N} (1-\mathcal{B}_i)}
\end{eqnarray}
where $\mathcal{B}$ is the object masks, $\mathcal{L}_{depth}^{fg}$ is the foreground depth loss and $\mathcal{L}_{depth}^{bg}$ is the background depth loss. 

The combined depth loss can be written as: 

\begin{equation}
    \mathcal{L}_{depth}=\lambda \cdot \mathcal{L}_{depth}^{fg} + (1-\lambda)\cdot \mathcal{L}_{depth}^{bg}
\end{equation}

\noindent where hyperparameter $\lambda$ is included to balance the effect on foreground and background learning. Lastly, the overall loss function is proposed as: 
\begin{equation}
	\mathcal{L} = \mathcal{L}_{disp} + \beta \cdot \mathcal{L}_{depth}
	\label{eq:final_loss}
\end{equation}
where hyperparameter $\beta$ is included to normalize the scales of disparity-based and depth-based loss functions. 

Although the ratio of foreground and background shown in Table~\ref{tab:data} suggests weighting of 0.8 for foreground and 0.2 for background ($\lambda=0.8$), our experimental results demonstrate otherwise. To investigate this phenomenon, extensive experiments were conducted to study the properties of depth-based loss function and effect of hyperparameter $\lambda$ on the overall performance of depth estimation, which will be discussed in the next section. 

\section{Experimental Details} \label{sec:experiment}
\subsection{Datasets}
\noindent \textbf{The KITTI 2015} stereo dataset contains images of natural scenes (city and rural areas and highways) collected in Karlsruhe, Germany. It contains 200 training stereo image pairs with sparse ground truth disparities, collected using LiDAR sensor; and 200 testing image pairs without ground truth disparities. KITTI allows performance evaluation by submitting final results to their evaluation server. Following~\cite{chang2018pyramid}, we perform hold-out validation by splitting the 200 training images into 160 for training and 40 for validation. All the results presented in Section \ref{sec:exp-results} are computed using the same validation set unless stated otherwise. 

\hfill

\noindent \textbf{The DrivingStereo} dataset is a large scale stereo dataset, covering a diverse set of driving scenarios and different weather conditions; containing over 174,437 stereo pairs for training and 7751 pairs for testing~\cite{yang2019drivingstereo}. Sparse ground truth disparities are provided for the training sets only. We use the DrivingStereo dataset to pre-train the stereo matching model, before fine-tuning on smaller KITTI dataset. Similarly, the dataset is split into training and validation set. Four subsets were randomly selected as the validation set while the remaining are used as the training set.

\hfill

\noindent \textbf{The Scene Flow} dataset is a large collection of synthetic stereo images with dense disparity ground truth. It comprises three subsets with different settings: FlyingThings3D, Driving and Monkaa. It consists of 35,454 training and 4,370 testing images. The size of each image is $960\times540$. As the maximum disparity in this dataset is larger than our pre-defined maximum disparity value, $D_{max}$, any pixel with disparity larger than the $D_{max}$ is neglected in the loss computation. This dataset is used to validate the effectiveness of the combined disparity-based and depth-based loss functions. The results demonstrate that the combined loss function can effectively improves the performance of depth estimation at all distances in any scenarios (real-world and synthetic scenes). 

\subsection{Metrics}
We evaluate the performance of disparity estimation by the proposed method, using endpoint-error~(EPE), and 3-pixel~(D1) metric as implemented by the KITTI benchmark. EPE computes the mean absolute error of the estimated disparity using the ground truth. The D1 metric on the other hand counts the ratio of pixels with EPE of $<3$ pixel or $<5\%$ based on the ground truth. 

We also evaluate the depth estimation accuracy for objects up to 80 meters from the camera. Predicted disparity (pixel) are converted into depth (metre). Evaluation is conducted by calculating the EPE for depth values. This metric provides valuable insights on the performance of depth estimation at different depth range.

\begin{table*}[ht]
\centering
\resizebox{0.65\textwidth}{!}{%
\begin{tabular}{c c|c c c c c c c c }
\hline
\multicolumn{2}{c|}{Loss Functions} & \multicolumn{8}{c}{KITTI~2015 Depth EPE (m)} \\ \cline{0-9} 
$\mathcal{L}_{disp}$ & $\mathcal{L}_{depth}$ & 0-10 & 10-20 & 20-30 & 30-40 & 40-50 & 50-60 & 60-70 & 70-80 \\ \hline
\checkmark &  & \textbf{0.12} & 0.38 & 0.89 & 1.50 & 2.42 & 3.69 & 5.39 & 6.47 \\ 
 & \checkmark & \textbf{0.12} & \textbf{0.35} & 0.87 & 1.51 & 2.48 & 3.68 & 5.12 & 5.89 \\ 
\checkmark & \checkmark & 0.13 & 0.40 & \textbf{0.86} & \textbf{1.46} & \textbf{2.33} & \textbf{3.53} & \textbf{4.61} & \textbf{5.66}  \\ \hline
\end{tabular}}
\caption{Accuracy of depth estimation at different depth intervals using different loss functions. The pixels are separated into bins according to their true depth values. The results illustrate that depth-based loss function can effectively improve the accuracy of depth estimation for both foreground and background that are located at far distances.}
\label{tab:loss_compare}
\end{table*}

\begin{table}[t]
\centering
\footnotesize
\begin{tabular}{l|c c c|c|c}
\hline
\multirow{3}{*}{Loss Functions}                            & \multicolumn{3}{c|}{KITTI 2015}                  & DS & SF  \\ \cline{2-6} 
                                                           & \multicolumn{3}{c|}{D1}                          & D1       & EPE      \\ \cline{2-6} 
                                                           & Fg             & Bg             & All            & All    & All        \\ \hline
$\mathcal{L}_{disp}$                                       & 1.89          & 1.83          & 1.98          & 0.68   &1.33       \\ 
$\mathcal{L}_{depth}$                                      & 1.66          & 1.85          & 1.90          & 0.58      &1.94    \\ 
$\mathcal{L}_{disp} \ \& \ \mathcal{L}_{depth}$            & 1.53          & 1.95          & 1.95          & 0.53         &1.10  \\ 
$\mathcal{L}_{disp} \ \& \ \lambda\mathcal{L}_{depth}^{(fg, bg)}$ & \textbf{1.31} & \textbf{1.80} & \textbf{1.78} & \textbf{0.48} &-  \\ \hline
\end{tabular}
\caption{Ablation study of different loss function combination. DS: DrivingStereo. SF: Scene Flow. The high error rate resulted when tested with Scene Flow training set trained using $\mathcal{L}_{depth}$ is because the test set consists of many floating objects that are located very close to the camera. As $\mathcal{L}_{depth}$ has the opposite property of $\mathcal{L}_{disp}$, worse result is expected for close distance pixels.}
\label{tab:loss_abla}
\end{table}

\subsection{Implementation details} 
The proposed loss functions are implemented in conjunction with network architecture proposed in PSMNet~\cite{chang2018pyramid}. The PSMNet is an effective 3D stereo matching network that is commonly used as backbone for disparity estimation~\cite{song2020adastereo, wang2019pseudo, you2019pseudo, zhang2019adaptive, yao2020content}. The network is implemented using PyTorch framework and is trained end-to-end with Adam $(\beta_1=0.9, \beta_2=0.999)$ optimizer. Our data processing is the same as PSMNet where the input images are normalized and randomly cropped to size $H=256$ and $W=512$. The maximum disparity is set to 192. All ground truth disparity beyond the maximum disparity, or below 0, are ignored in our experiments. The hyperparameter $\beta$ is set to $1.0$ for all experiments. We trained the model from scratch using the DrivingStereo dataset with a constant learning rate of 0.001 for 10 epochs. The same model was also trained using the Scene Flow dataset to study the effect of disparity-based and depth-based loss function. Similarly, the training was conducted for 10 epochs using constant learning rate of 0.001. 

We then used the pre-trained model (with DrivingStereo dataset) and finetuned on KITTI training set for 300 epochs. The learning rate for finetuning process starts at 0.001 and is decreased to 0.0001 after 200 epochs. Following~\cite{chang2018pyramid}, the finetuning process is prolonged to 1000 epochs with learning rate begins at 0.001 and decreased to 0.0001 after $\frac{2}{3}$ of total epochs before submission to KITTI evaluation server. The batch size is set to 12 for training on 2 NVIDIA RTX 6000 Quadro GPUs.

\subsection{Experimental Results and Ablation Study}
\label{sec:exp-results}
To validate the effectiveness of each component proposed in this work, several experiments with different loss function combinations are conducted using KITTI 2015 validation set as well as DrivingStereo and Scene Flow testing sets. 

\begin{table}[t]
\centering
\resizebox{0.48\textwidth}{!}{%
\begin{tabular}{ccc|ccc|ccc}
\hline
\multicolumn{3}{c|}{\multirow{2}{*}{Loss Functions}} & 
\multicolumn{6}{c}{KITTI 2015} \\ \cline{4-9}
\multicolumn{3}{c|}{} &
\multicolumn{3}{c|}{D1}  &
\multicolumn{3}{c}{EPE}  \\ \hline
$\mathcal{L}_{disp}$ & $\mathcal{L}_{depth}^{fg}$ & $\mathcal{L}_{depth}^{bg}$ & All & Fg & Bg & All & Fg & Bg            \\ \hline
\checkmark           &                            &                            & 1.98         & 1.89         & 1.83          & 0.65          & 1.18  & 0.53                  \\ 
\checkmark           & \checkmark                 &                            & 2.03         & 1.69          & 1.99          & 0.68          & 1.16          & 0.58                     \\ 
\checkmark           &                            & \checkmark                 & 2.06          & 2.04          & 1.87          & 0.65          & 1.18          & 0.53                   \\ 
\checkmark           & \checkmark                 & \checkmark                 & \textbf{1.78} & \textbf{1.31} & \textbf{1.80} & \textbf{0.63} & \textbf{1.11} & \textbf{0.53}   \\ \hline
\end{tabular}
}
\caption{Ablation study of the proposed foreground and background depth-based loss functions. Depth-based loss function provides better improvement for foreground pixels than background pixels, but when the foreground and background depth-based loss functions are weighted properly, superior results can be obtained.}
\label{tab:obj_loss_abla}
\end{table}

\begin{figure*}[ht]
\centering
	\includegraphics[width=.3\textwidth]{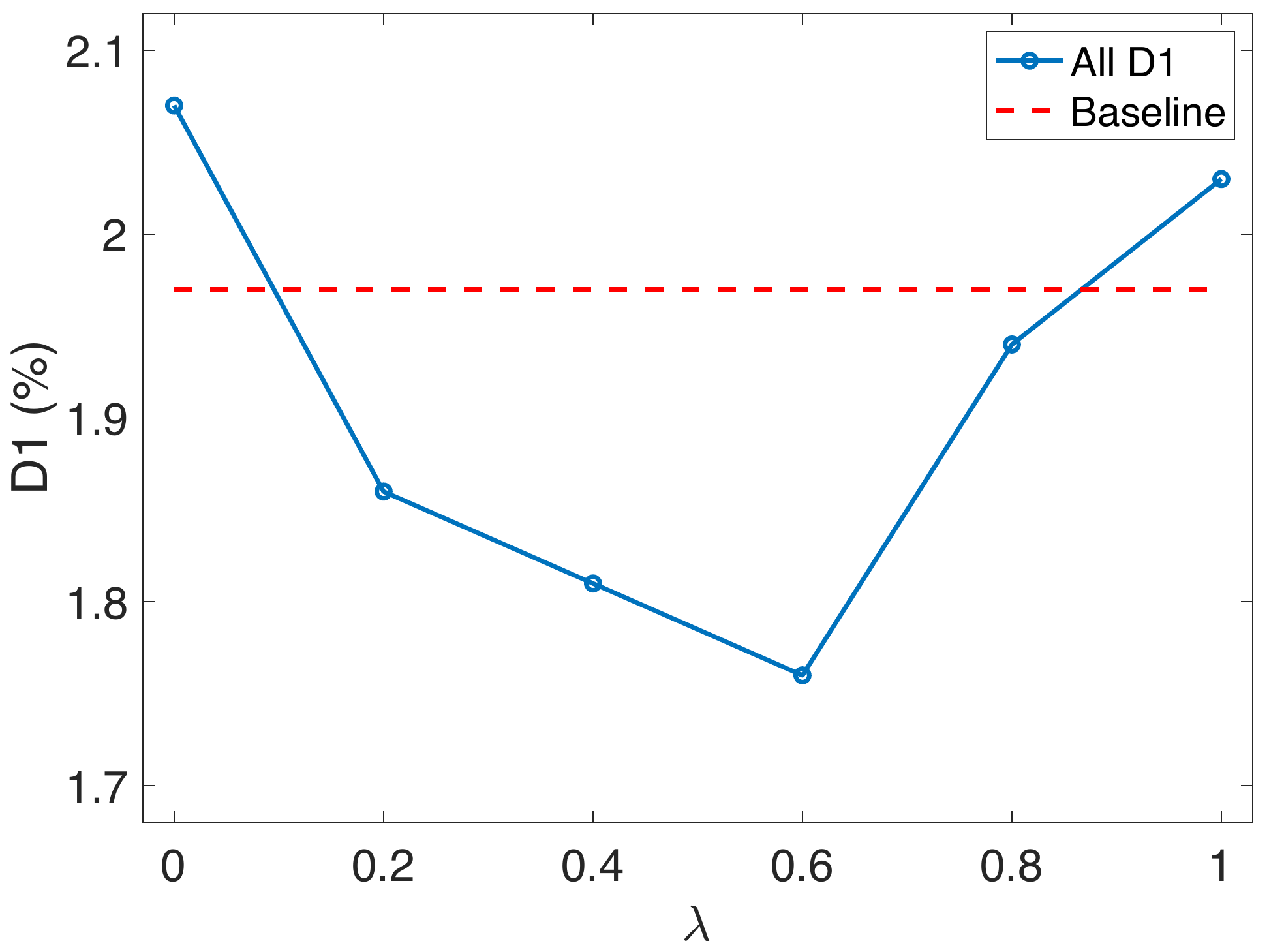}
	\includegraphics[width=.3\textwidth]{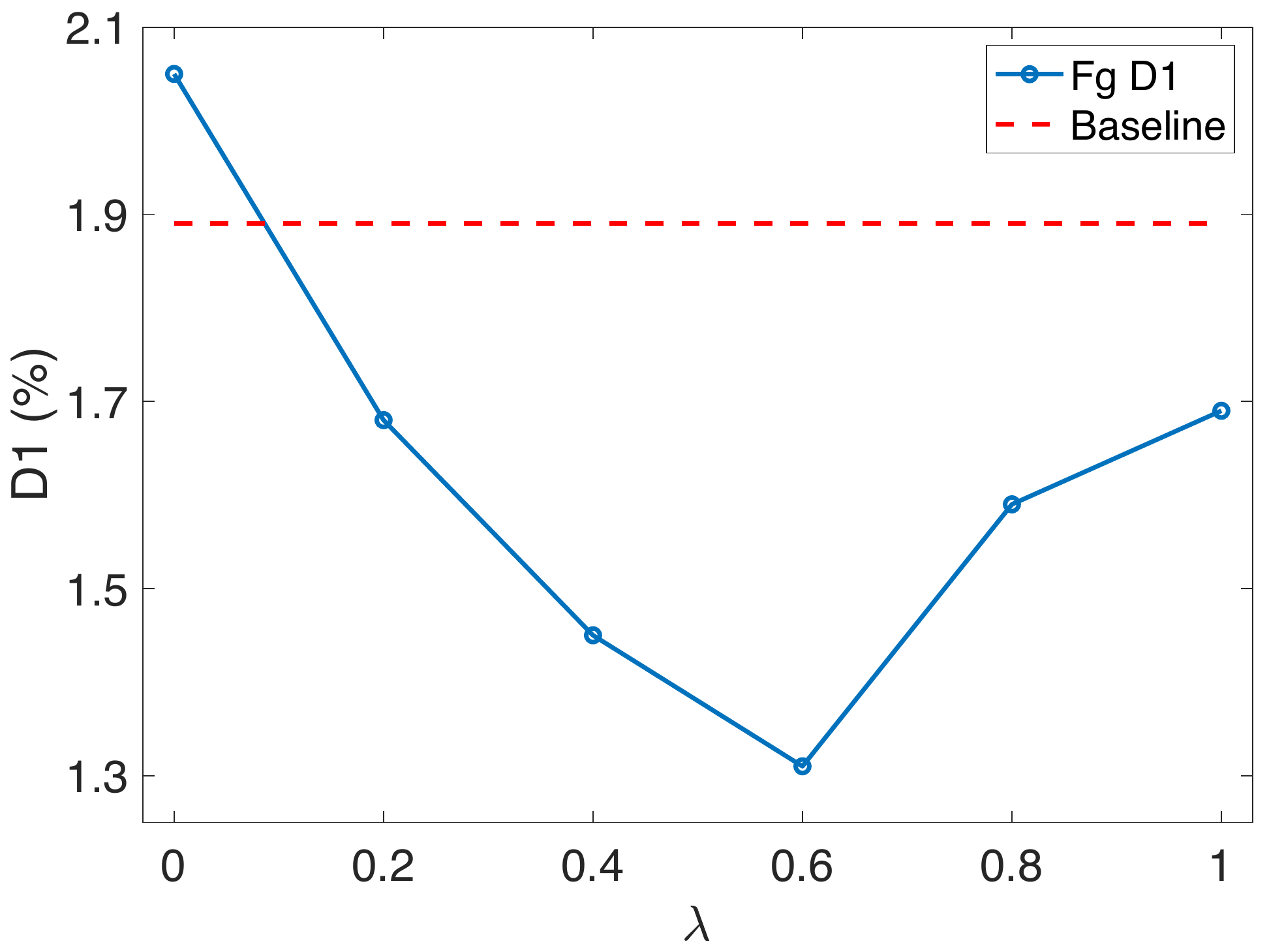}
	\includegraphics[width=.3\textwidth]{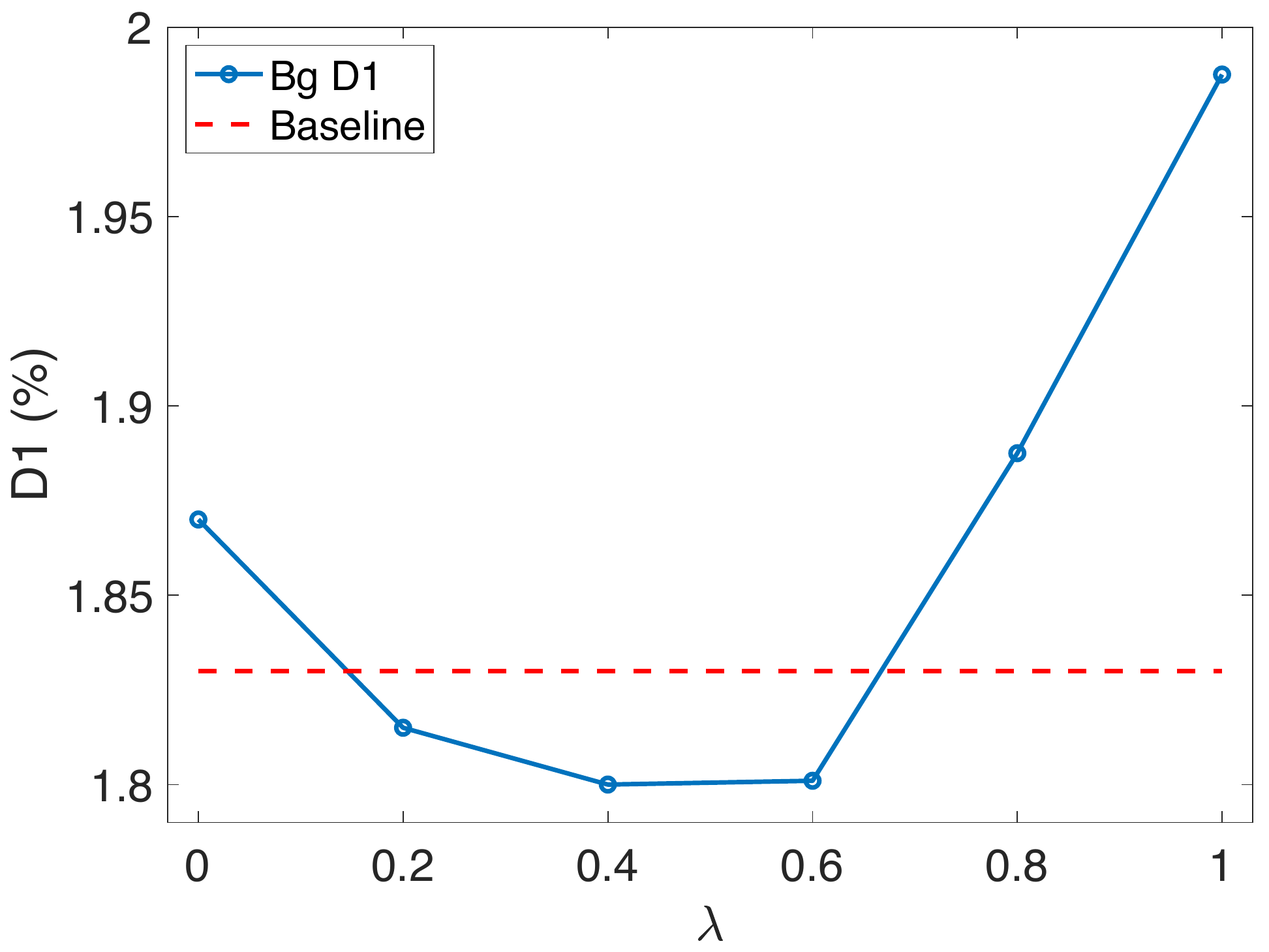}
	\caption{These graphs show the relationship between the balancing term $\lambda$ and the D1 error of all pixels~(left), foreground~(middle) and background~(right). They also show that $\lambda = 0.6$ yields the best performance. Red dotted line indicates the performance of the baseline method (PSMNet).}
	\label{fig:lambda}
\end{figure*}

\noindent \textbf{Ablation study for disparity and depth loss functions:}~ We seek to investigate the regularization property of the depth-based loss function, and how it impacts the performance of our trained stereo matching network. We do so by conducting four sets of experiments using different loss functions. Specifically, we trained the network with (1) disparity loss function only, (2) depth loss function only, (3) disparity and depth loss functions and (4) disparity and the proposed weighted foreground/background depth loss functions. Note that in our experiments, the predicted disparity and its ground truth are converted into depth and the depth loss is computed by Eq.~(\ref{eq:depthLossFunction}). 

Depth-based loss function is added to regularize the training, aiming to mitigate the over-fitting caused by the disparity-based loss function, by allowing greater training signals for the objects positioned at further distances. The results of these experiments are shown in Table~\ref{tab:loss_compare} and Table~\ref{tab:loss_abla}. In Table~\ref{tab:loss_compare}, the EPE metric is selected to study the accuracy of different loss functions for depth estimation at different distance ranges. As shown, training using depth-based loss function achieved better accuracy for objects located at greater distances ($\geq10m$) compared to training using disparity-based loss function. 

Also, by combining the two loss functions, the network achieves even better accuracy for objects located beyond 20m. Although the performance of the close distance pixels ($0-20$m) have deteriorated slightly (around $1\%$), this is a relatively small price to pay for significant improvement in the accuracy of long range measurements. It is also interesting to note that depth-based loss function also improves the accuracy of measurements for foreground objects irrespective of their depth. Table~\ref{tab:loss_abla} shows that by including the depth-based loss function, the overall accuracy for foreground objects are improved by $19\%$.  

\begin{figure}
\centering
	\subfloat[]{\includegraphics[width=.23\textwidth]{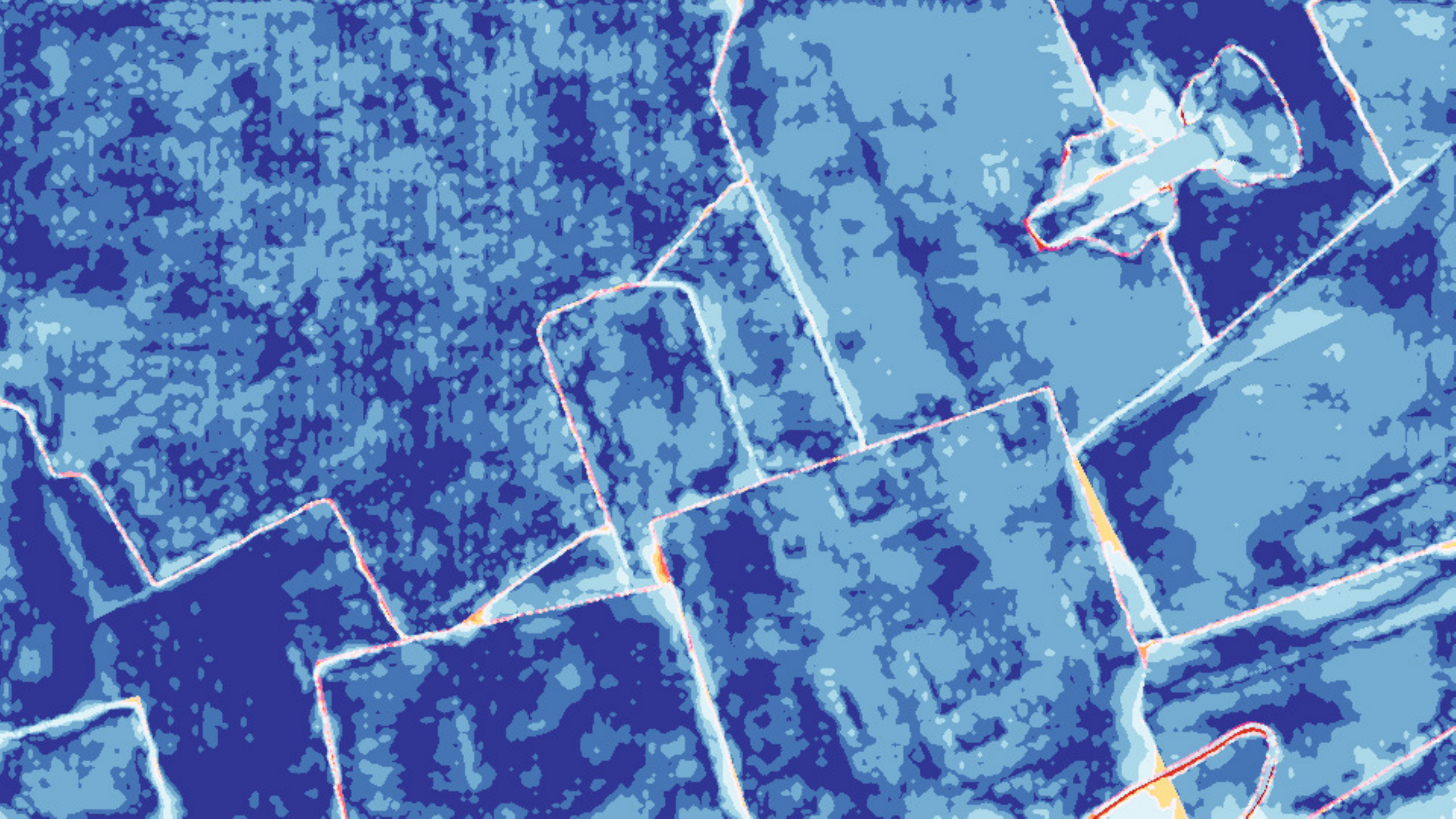}}
	\hspace{0.1mm}
	\subfloat[]{\includegraphics[width=.23\textwidth]{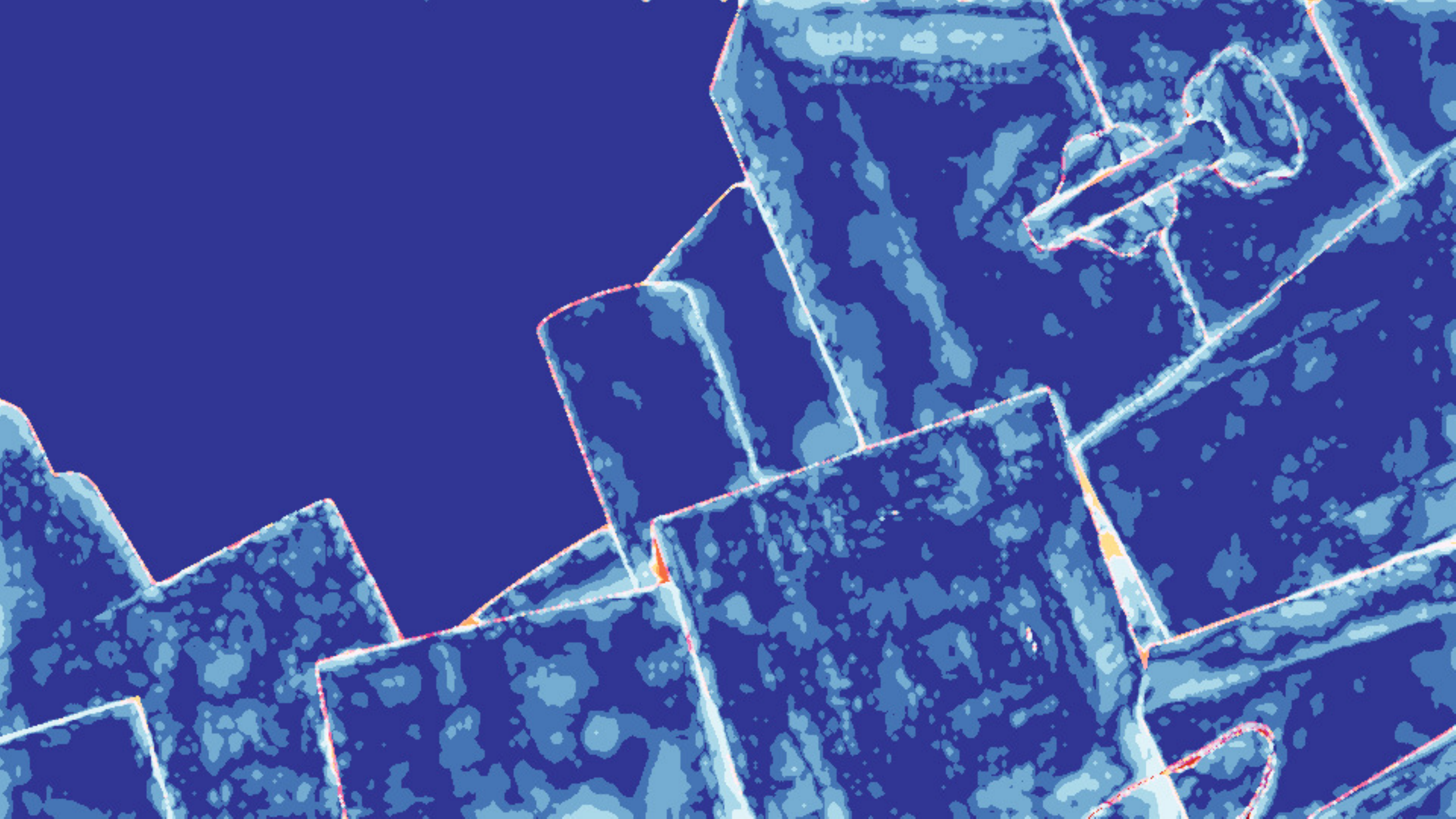}}

	\caption{Qualitative results of Scene Flow dataset comparing the performance of disparity-based loss function and the combination of disparity-based and depth-based loss function. The error map of the former is shown in (a) and the latter is shown in (b). By including depth-based loss function, superior results can be obtained especially for the pixels located at very far distances.}
	\label{fig:sf_results}
\end{figure}

We further demonstrate the effectiveness of depth-based loss function using the Scene Flow dataset. As illustrated in Fig.~\ref{fig:sf_results}, by including the depth-based loss function, we have consistently achieved incredibly low errors for distant foreground and background. The background of the image included in Fig.~\ref{fig:sf_results} has true disparity values ranging between $[1.1, 1.6]$ pixels.

\begin{table}[t]
\centering
\resizebox{0.48\textwidth}{!}{%
\begin{tabular}{l|cccccccc}
\hline
\multirow{2}{*}{Methods}                        & \multicolumn{8}{c}{Range (m)}                                                                                                \\ \cline{2-9} 
                                                & 0-10          & 10-20         & 20-30         & 30-40         & 40-50         & 50-60         & 60-70         & 70-80         \\ \hline
PSMNet~\cite{chang2018pyramid} & 0.18 & 0.36          & 0.97          & 2.02          & 2.94          & 4.61          & 6.03          & -             \\ 
SDN~\cite{you2019pseudo}       & 0.21          & \textbf{0.35} & 0.87 & 1.80 & 2.67 & 4.27 & 5.82 & -             \\
SDN+GDC~\cite{you2019pseudo}       & 0.21          & \textbf{0.35} & \textbf{0.84} & 1.74 & 2.59 & 4.14 & 5.72 & -             \\   \hline
LR-PSMNet                                       & \textbf{0.11} & \textbf{0.35} & \textbf{0.84} & \textbf{1.44} & \textbf{2.33} & \textbf{3.29} & \textbf{4.67} & \textbf{5.53} \\ \hline
\end{tabular}%
}
\caption{Mean depth error~(m) of KITTI~2015 validation set over various depth range. Our approach resulted significant improvement for very far-away pixels without sacrificing the accuracy of close distance pixels.}
\label{tab:mean_depth_err}
\end{table}

\vfill 
\noindent \textbf{Ablation study for foreground and background depth loss functions:}~We tackle the imbalance between foreground and background by designing a pair of novel depth-based loss functions, namely foreground specific $\mathcal{L}^{fg}_{depth}$ and background specific $\mathcal{L}^{bg}_{depth}$, that are appropriately weighted. The ratio between the foreground and background data, listed in Table~\ref{tab:data}, suggests a weighting of 0.8 for foreground and 0.2 for background ($\lambda\,=\,0.8$). However, from our experiments, we have found that depth-based loss function has better performance for foreground than background pixels. Therefore, by giving less weights to the foreground pixels and more to background ones, we may achieve a better balance. This is explained in the next subsection. 

To support our argument, we have conducted two additional experiments using the disparity-based loss function with either foreground specific ($\lambda\,=\,1$) or background specific ($\lambda\,=\,0$) components. The results are tabulated in Table~\ref{tab:obj_loss_abla}. Within expectation, the results demonstrated that $\mathcal{L}^{fg}_{depth}$ is advantageous for foreground prediction. However, solely including $\mathcal{L}^{bg}_{depth}$ worsens the accuracy for background objects as well as the overall accuracy. Regardless, $\mathcal{L}^{bg}_{depth}$ is still required to improve the accuracy of background objects located at far distances. As such, both depth-based loss components are needed to improve the overall accuracy at all distances. 

\noindent \textbf{Analysis of balancing term $\lambda$:}~Hyperparameter $\lambda$ balances the contributions of foreground specific $\mathcal{L}^{fg}_{depth}$ and background specific $\mathcal{L}^{bg}_{depth}$ components to the total loss. We selected the optimal value for $\lambda$ using grid-search between 0 and 1 with interval of 0.2. As it was mentioned earlier, the ratio between foreground and background data in the KITTI 2015 dataset implies the $\lambda$ to be $0.8$ for optimal performance. However, we observed that the overall, foreground and background error curves are similar 'V' shape curves and the optimal results for $\lambda$ is close to 0.6. 

Fig.~\ref{fig:lambda} shows that including the $\mathcal{L}^{fg}_{depth}$ in loss calculations (by setting $\lambda>0$) lowers the D1 error for foreground objects. However, the effect of including $\mathcal{L}^{bg}_{depth}$ is less pronounced. 
Our conjecture is that the values of $\mathcal{L}^{fg}_{depth}$ and $\mathcal{L}^{bg}_{depth}$ somehow reinforce each other and by including both terms, the network produces better accuracy for background despite that the depth-based loss function by itself does not perform well for background depth measurement. However, when the $\mathcal{L}^{bg}_{depth}$ weight is reduced (say for $\lambda\geq0.8$), the performance deteriorates quickly. 

\begin{table}[t]
\centering
\resizebox{0.47\textwidth}{!}{%
\begin{tabular}{l|ccc|ccc}
\hline
\multirow{2}{*}{Method} & \multicolumn{3}{c|}{D1-All (\%)}                                                       & \multicolumn{3}{c}{D1-Noc(\%)} \\ \cline{2-7}
                        & \multicolumn{1}{c}{Bg} & \multicolumn{1}{c}{Fg} & \multicolumn{1}{c|}{all} & \multicolumn{1}{c}{Bg} & \multicolumn{1}{c}{Fg} & \multicolumn{1}{c}{all}               \\ \hline
AANet~\cite{xu2020aanet} & 1.99 & 5.39 & 2.55 & 1.80 & 4.93 & 2.32 \\
PSMNet~\cite{chang2018pyramid} & 1.86 & 4.62 & 2.32 & 1.71 & 4.31 & 2.14\\
SegStereo~\cite{yang2018segstereo} & 1.88 & 4.07 & 2.25 & 1.76 &3.70 & 2.08\\
DeepPruner~\cite{duggal2019deeppruner} & 1.87 & \textbf{3.56} & 2.15 & 1.71 & \textbf{3.18} & 1.95\\
SSPCVNet~\cite{wu2019semantic} & 1.75 & 3.89 & 2.11 & 1.61 & 3.40 & \textbf{1.91} \\ 
GwcNet~\cite{guo2019group} & 1.74 & 3.93 & 2.11 & 1.61 & 3.49 & 1.92 \\ \hline 
LR-PSMNet & \textbf{1.65} & 4.13 & \textbf{2.06} & \textbf{1.52} & 3.98 & 1.92\\ \hline
\end{tabular}
}
\caption{Benchmark results on KITTI 2015 test sets. All included results are obtained from the official KITTI 2015 benchmark.}
\label{tab:kitti}
\end{table}

\begin{figure*}[t]
\centering
    \includegraphics[width=.98\textwidth]{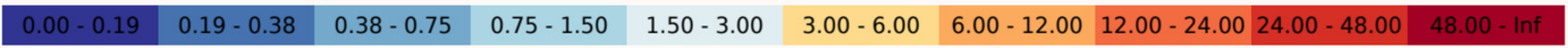}
	\includegraphics[width=.24\textwidth]{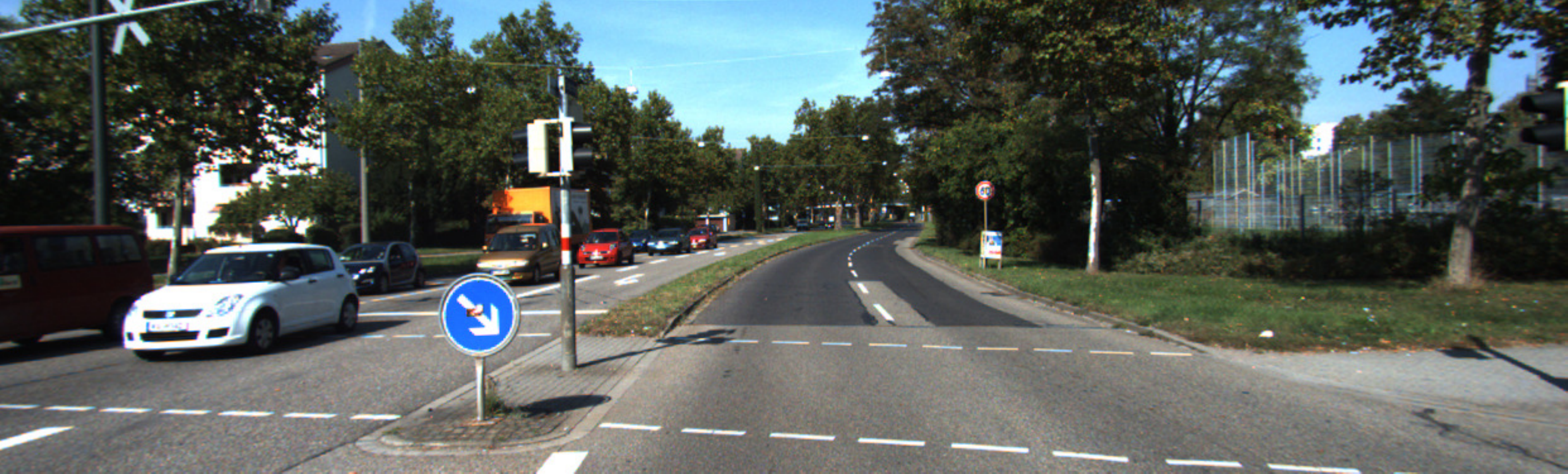}
	\includegraphics[width=.24\textwidth]{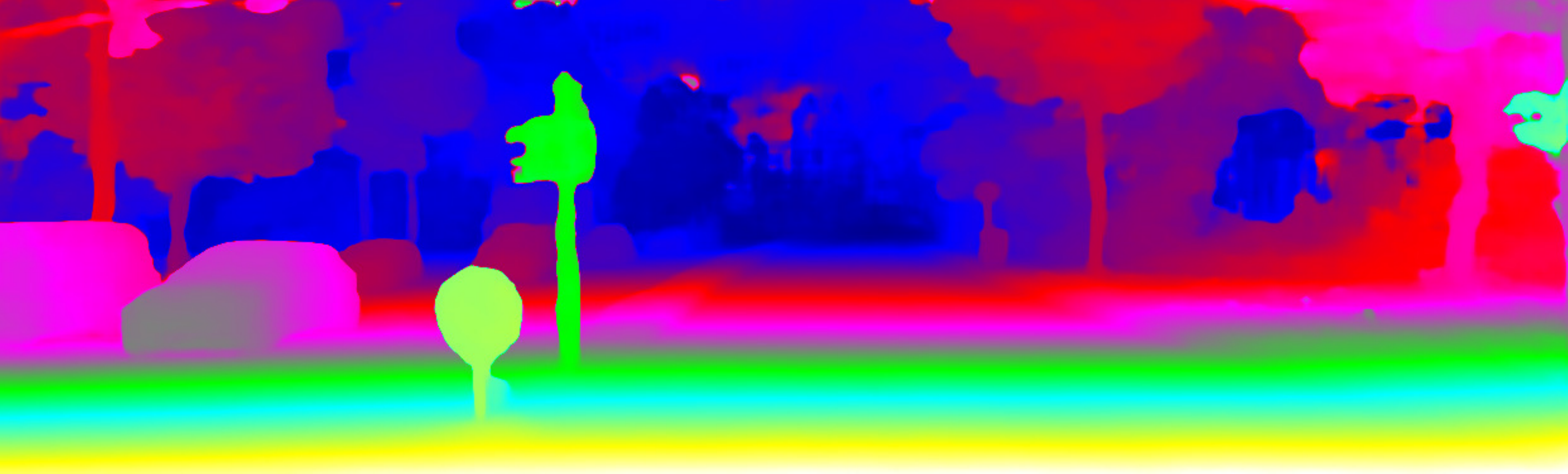}
	\includegraphics[width=.24\textwidth]{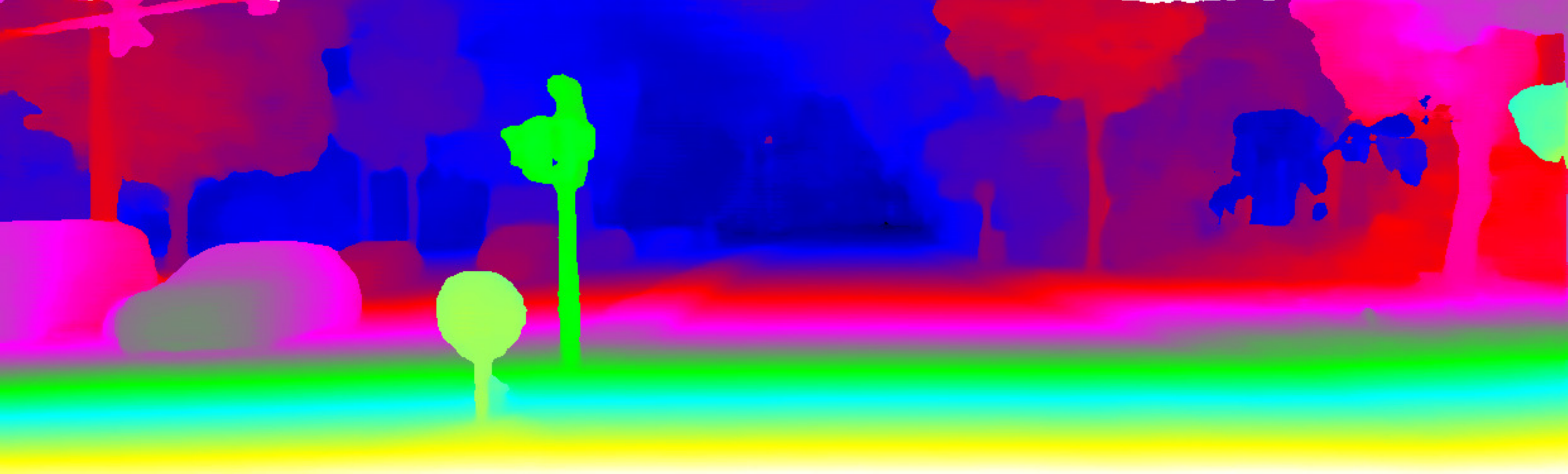}	
	\includegraphics[width=.24\textwidth]{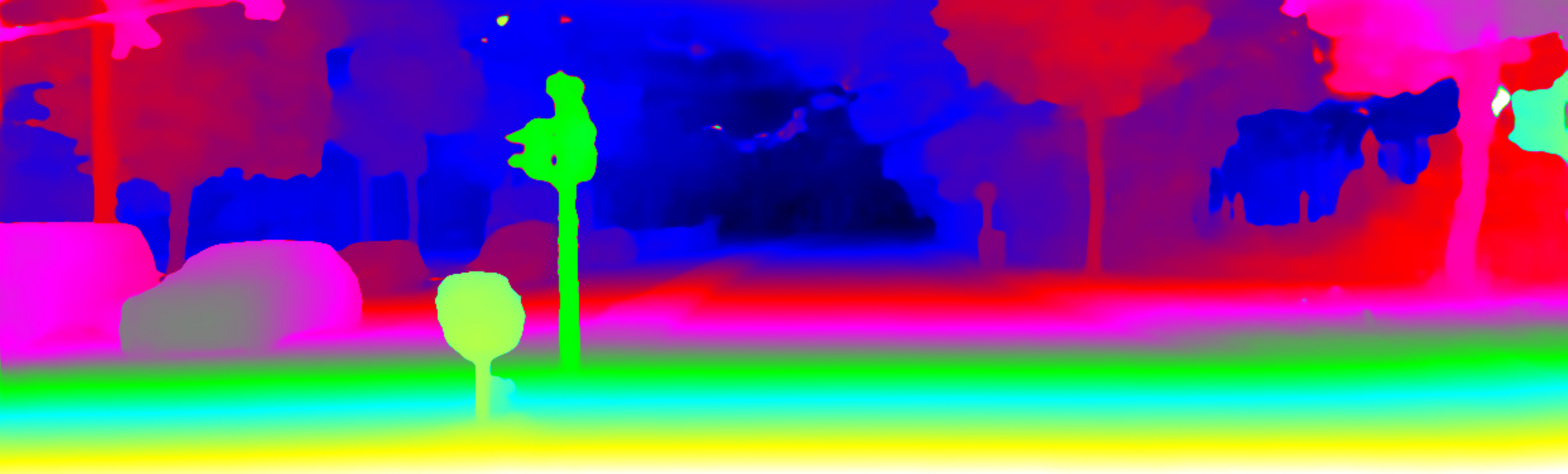}
	
	\vspace{-3.5mm}
    \subfloat[Left Image]{\includegraphics[width=.24\textwidth]{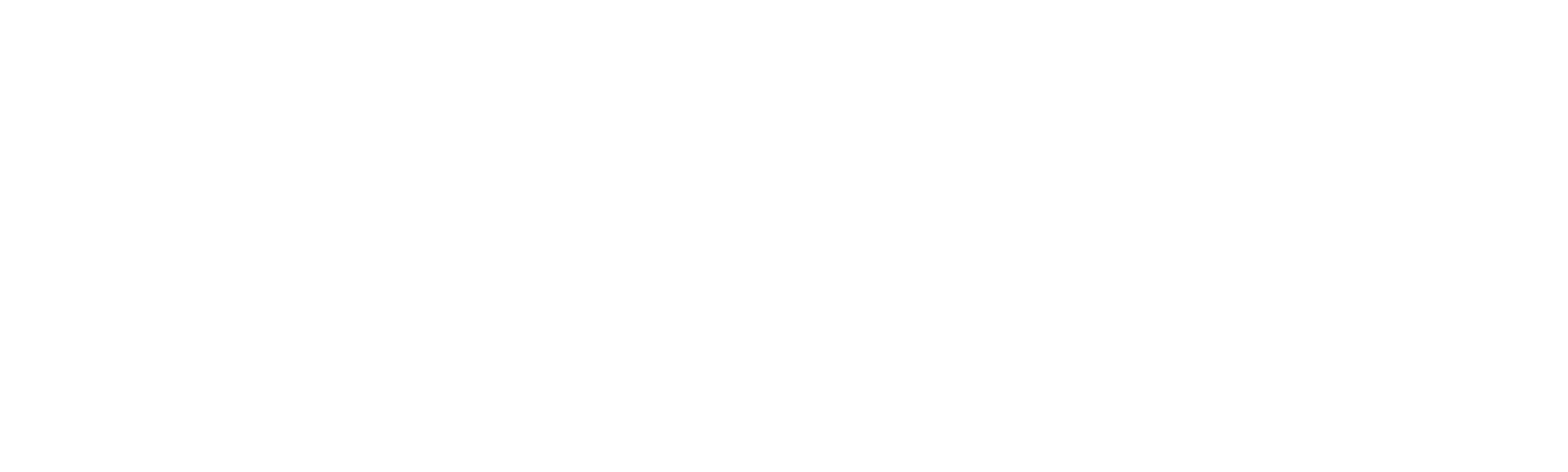}}
    \hspace{0.1mm}
    \subfloat[LR-PSMNet]{\includegraphics[width=.24\textwidth]{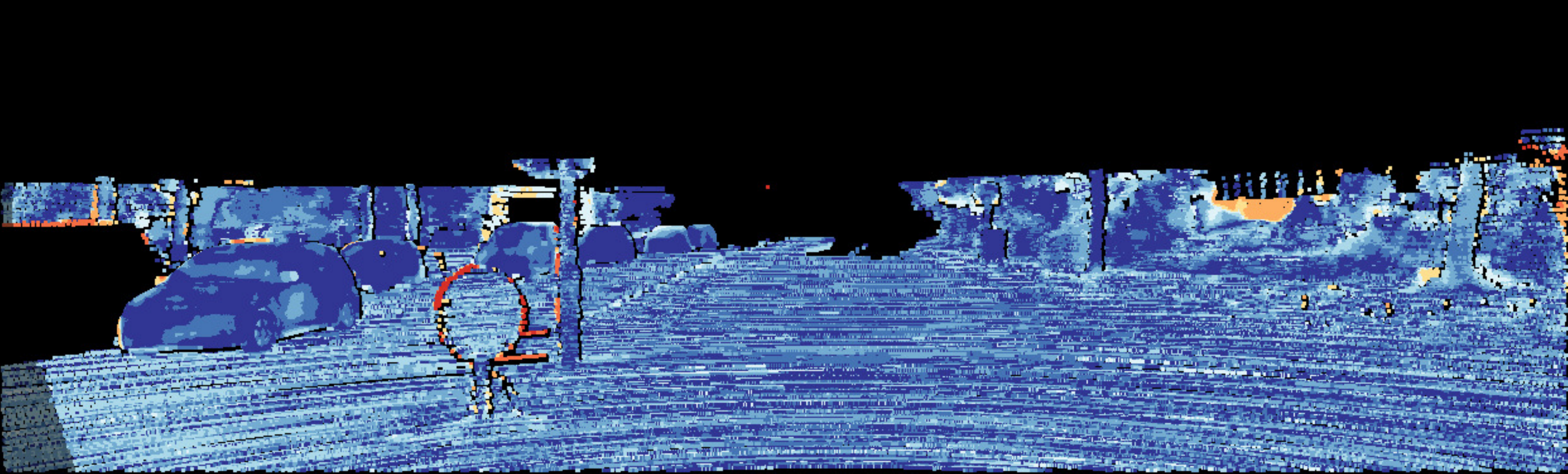}}
    \hspace{0.05mm}
    \subfloat[AcfNet]{\includegraphics[width=.24\textwidth]{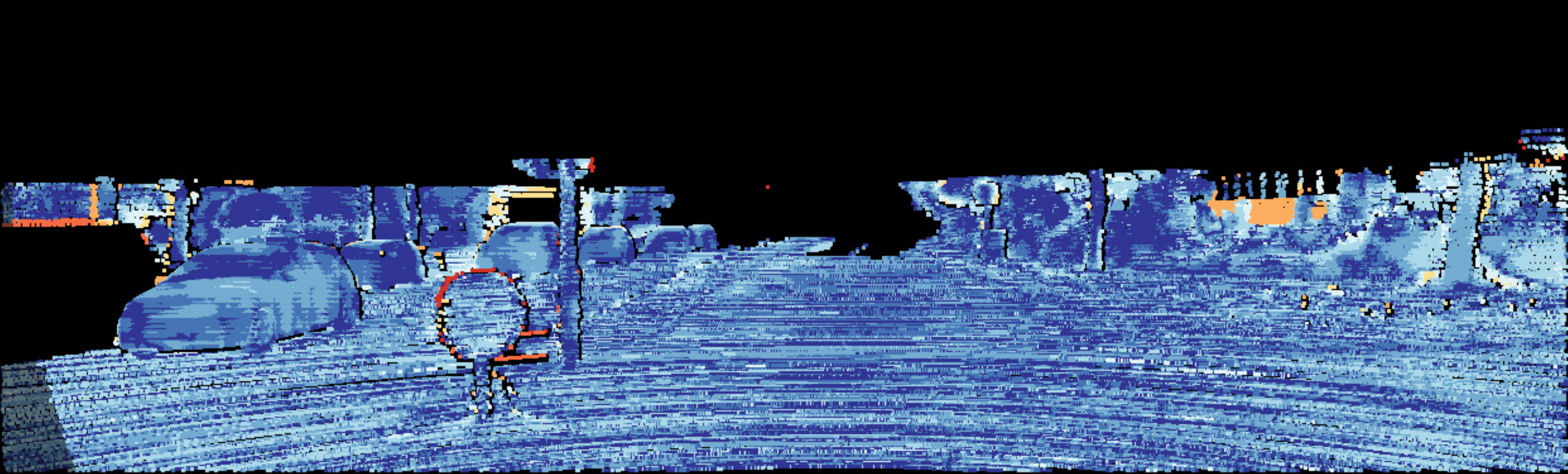}}
    \hspace{0.05mm}
	\subfloat[GANet]{\includegraphics[width=.24\textwidth]{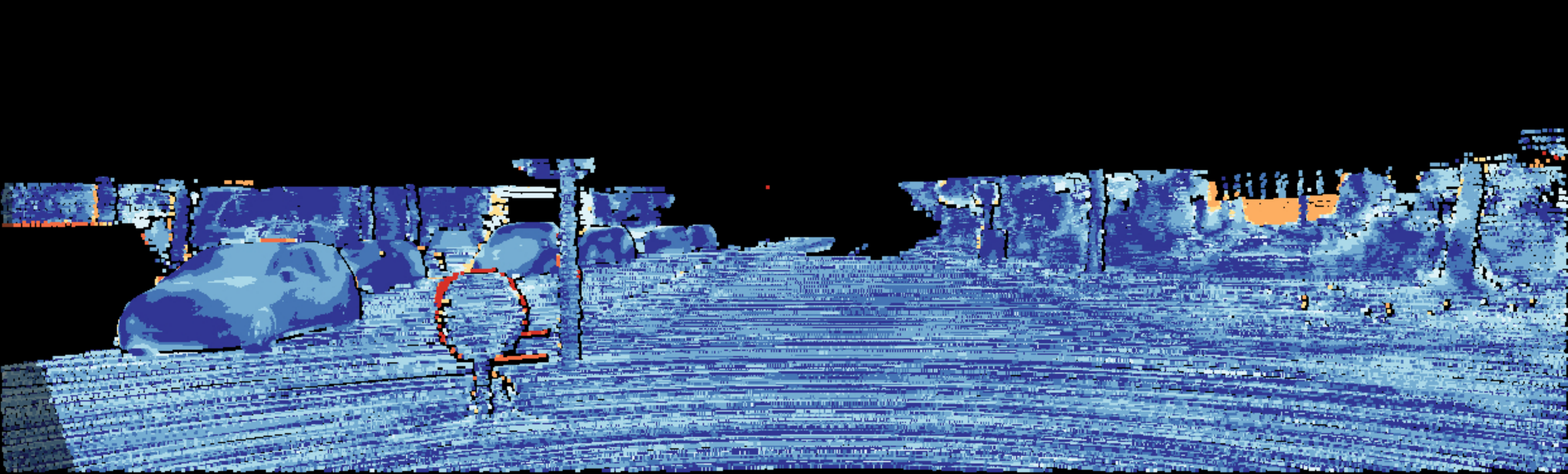}}
		
	\caption{Visualization results on KITTI 2015 dataset comparing our results with AcfNet~\cite{zhang2019adaptive} and GANet~\cite{zhang2019ga}. The numerical scale for color mapped on the error maps is provided on top of the figure.}
	\label{fig:kitti_results}
\end{figure*}

\begin{figure}[t]
\centering
	\includegraphics[width=.23\textwidth]{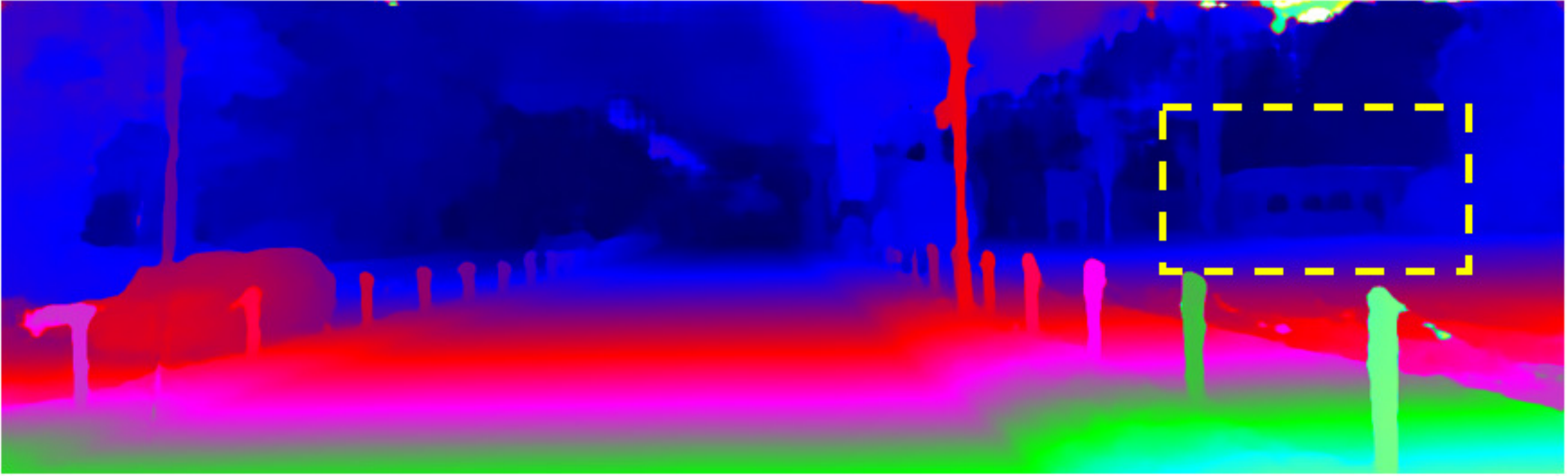}
	\includegraphics[width=.23\textwidth]{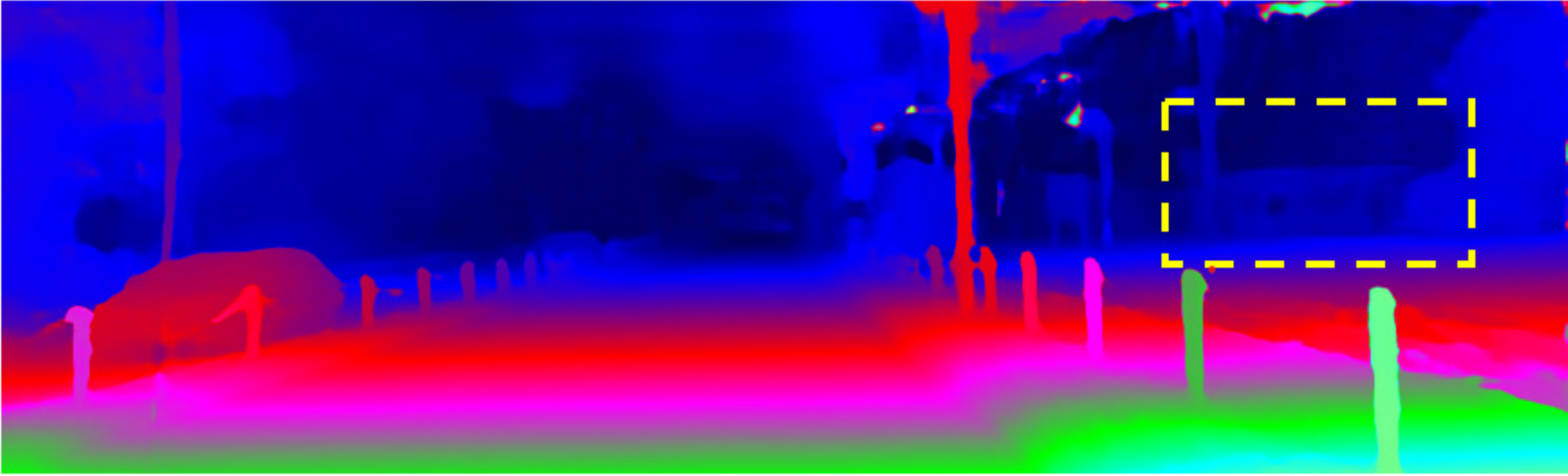}	
		
	\includegraphics[width=.23\textwidth]{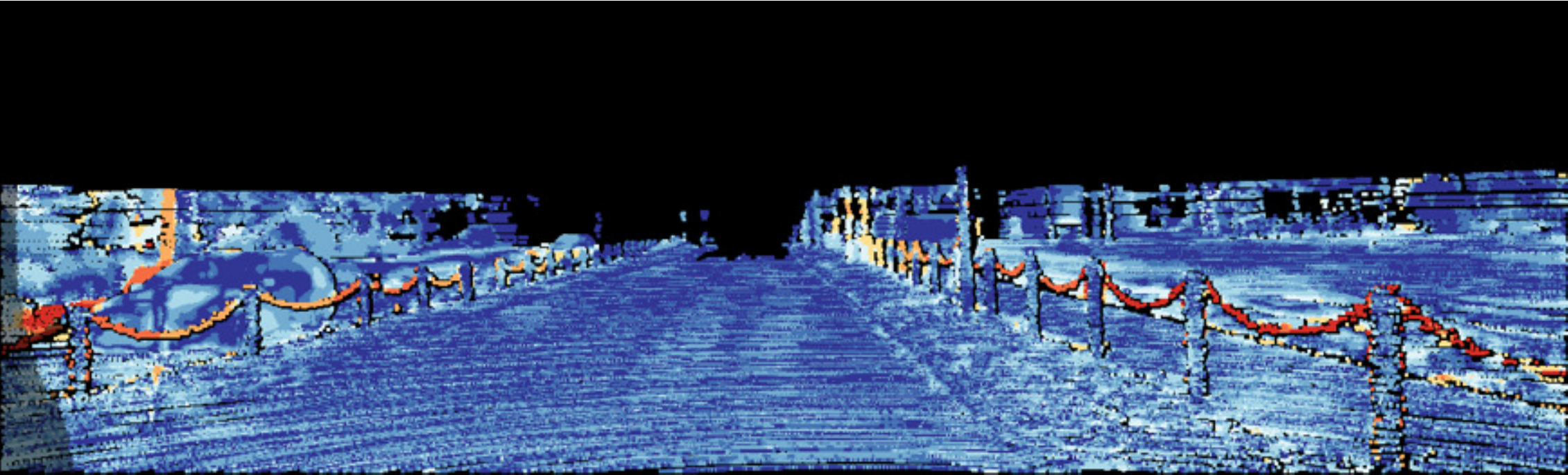}
	\includegraphics[width=.23\textwidth]{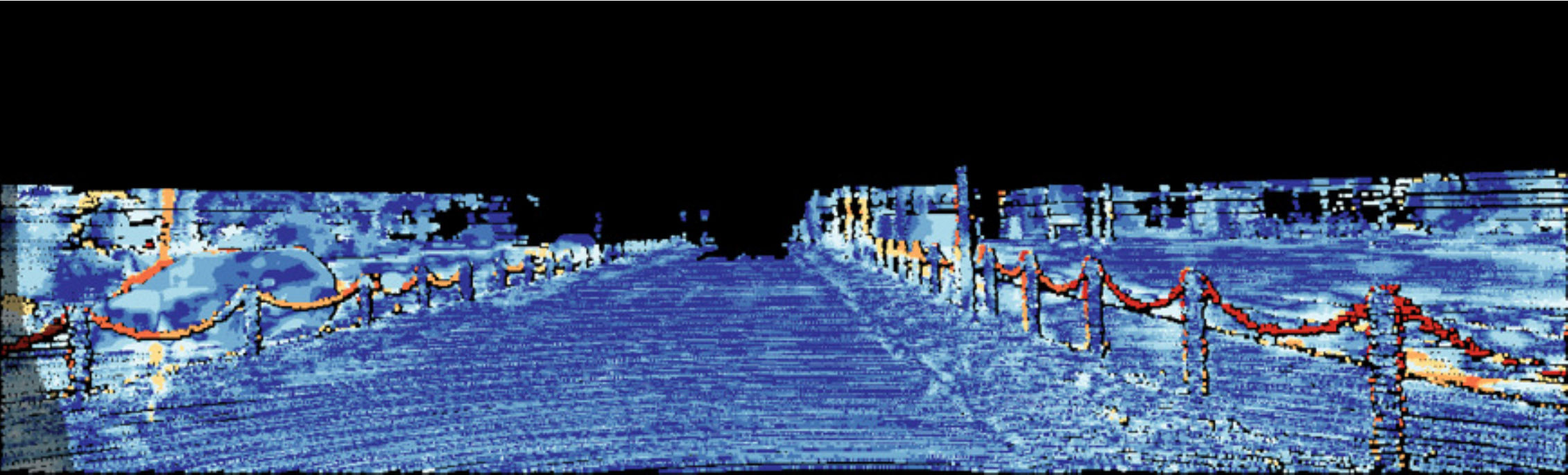}
	
	\includegraphics[width=.23\textwidth]{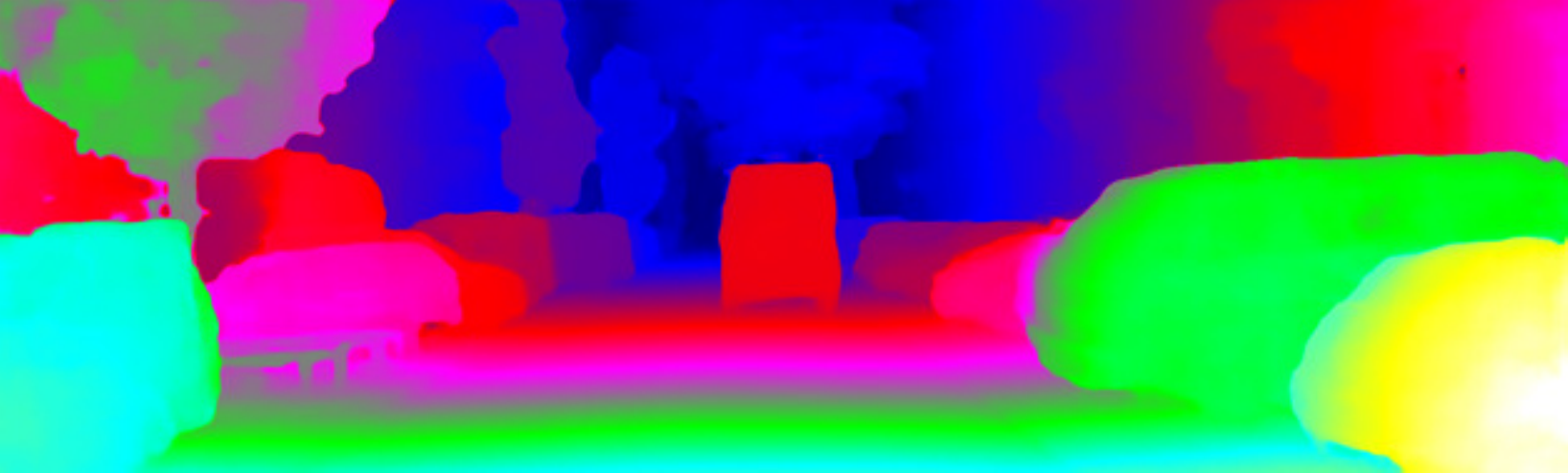}
	\includegraphics[width=.23\textwidth]{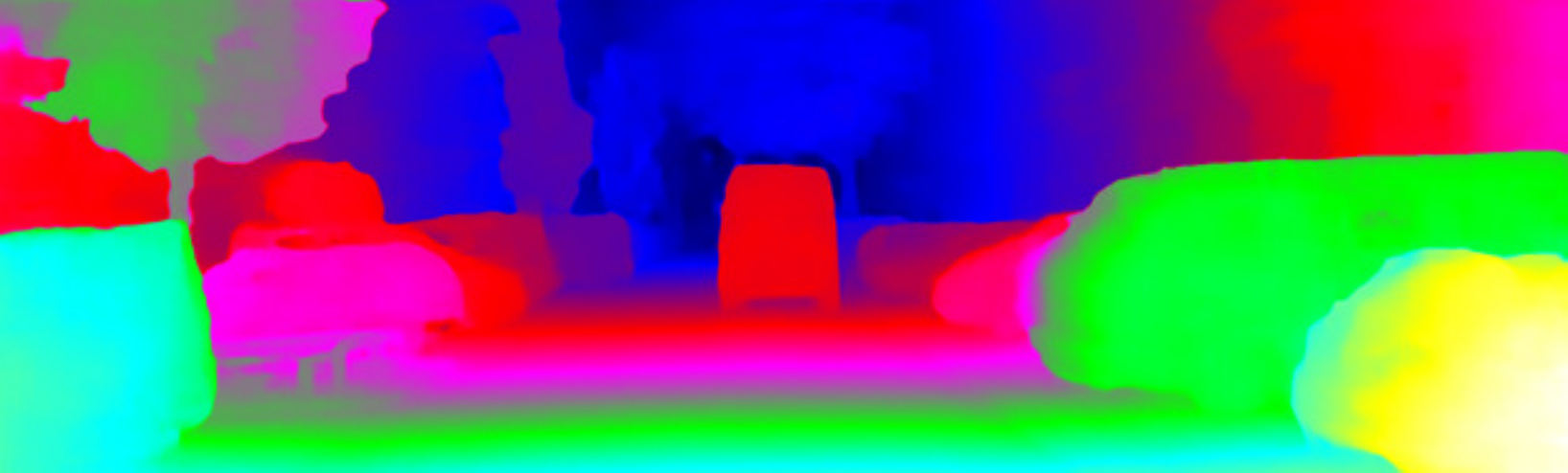}
	
	\vspace{-3.5mm}
	\subfloat[LR-PSMNet]{\includegraphics[width=.23\textwidth]{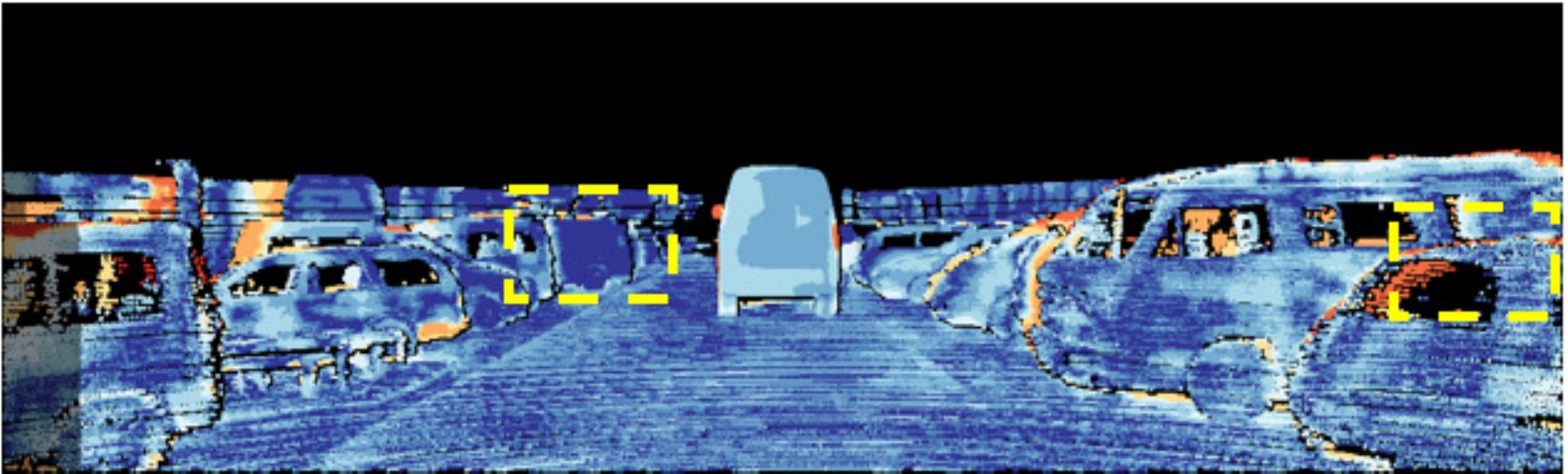}}
	\hspace{0.05mm}
	\subfloat[PSMNet]{\includegraphics[width=.23\textwidth]{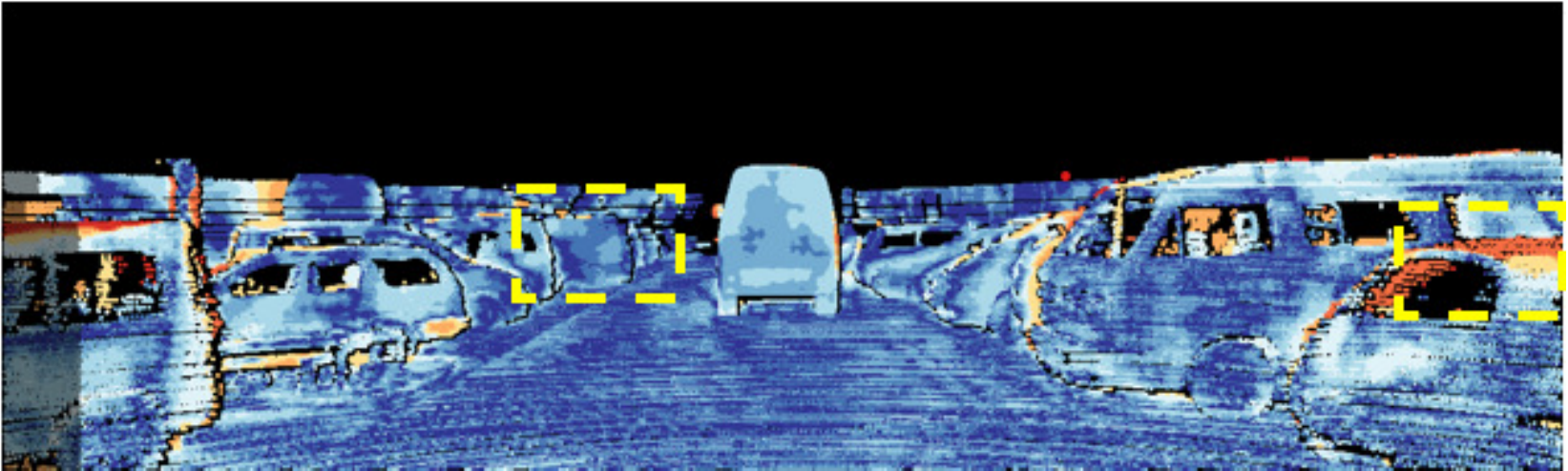}}

	\caption{Visualization of improvement over baseline method (PSMNet) on KITTI 2015 dataset. For each example, predicted disparity map is illustrated on top row and error map on the bottom row. Improved areas are highlighted with yellow dashed box.}
	\label{fig:kitti_improve}
\end{figure}

\noindent \textbf{Performance analysis of long range depth estimation:}~In this section, we compare the performance of long range depth estimation between the proposed method and the current state-of-the-art technique~(SDN~\cite{you2019pseudo}). We named our method LR-PSMNet~(LR: \underline{L}ong \underline{R}ange). As listed in Table~\ref{tab:mean_depth_err}, LR-PSMNet significantly improves the performance of PSMNet at all depth ranges, especially for estimates with true depth beyond 50m. Compared to SDN, LR-PSMNet does not suffer from exacerbation in performance for close-by estimates as our loss function combination is able to achieve good generalization at all depth ranges. More importantly, our results are on par with the SDN\,+\,GDC method that uses sparse but accurate depth information, measured by 4-beam LiDAR, to refine the depth estimates~\cite{you2019pseudo}.

\noindent \textbf{KITTI 2015 leaderboard:}~Although our work focuses on long range depth estimation, we also subjected the proposed method to KITTI performance evaluation exercise. The overall results of LR-PSMNet was $2.06\%$ as listed in Table \ref{tab:kitti}. This shows that by carefully redesign the loss function, the rank of the PSMNet ~\cite{chang2018pyramid} method is improved from rank 116 to 64 (recorded on $8^{\mathrm{th}}$ of November 2020). We argue that the higher error in foreground pixels is largely due to the limitation of the designed network as PSMNet, despite being a competitive method, has one of the highest foreground error among all state-of-the-art methods. Although the extension of our approach to other methods is straightforward; as it is, the proposed approach achieves comparable disparity estimation accuracy for foreground and background as compared to other high performing methods~\cite{zhang2019adaptive, zhang2019ga} - see Fig. \ref{fig:kitti_results} for details.

\section{Conclusion} \label{sec:conclusion}
This work simultaneously provides insights on (and solution to) the significant issues associated with bias in both dataset composition and disparity loss calculation for long range stereo depth estimation. We provided an effective solution by including foreground and background specific depth-based loss functions. We also showed that by separating the loss calculation for foreground and background and weighting those properly to balance the bias, we can improve the passive stereo depth calculation for long range foreground objects. Our experimental results demonstrated that the proposed method outperforms the state-of-the-art methods for long range stereo depth estimation. 

{\small
\bibliographystyle{ieee_fullname}
\bibliography{cvpr}
}

\end{document}